\documentclass{article}
\usepackage{arxiv}
\usepackage[utf8]{inputenc} 
\usepackage[T1]{fontenc}    
\usepackage{hyperref}  
\usepackage{url}          
\usepackage{booktabs}     
\usepackage{amsfonts}      
\usepackage{nicefrac}  
\usepackage{microtype}     	
\usepackage{amsmath}
\usepackage{graphicx}
\usepackage[numbers]{natbib}
\usepackage{doi}
\usepackage{lineno}
\usepackage{physics}
\usepackage{color}
\usepackage[dvipsnames]{xcolor}
\usepackage{caption}
\usepackage{subcaption}
\usepackage{xargs}
\usepackage{algorithm}
\usepackage{algpseudocode}
\usepackage{xcolor}
\renewcommand{\headeright}{}
\renewcommand{\undertitle}{}

\title{PDE-DKL: PDE-constrained deep kernel learning in high dimensionality}

\author{ Weihao ~Yan* \\
	Mathematics of Imaging \& AI\\
	University of Twente, The Netherlands \\
	\texttt{w.yan@utwente.nl} \\
        \And
	\href{https://orcid.org/0000-0003-0145-5069}{\includegraphics[scale=0.06]{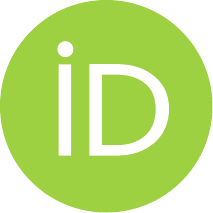}\hspace{1mm}Christoph ~Brune} \\
	Mathematics of Imaging \& AI\\
	University of Twente, The Netherlands \\
	\texttt{c.brune@utwente.nl} \\
	\And
	\href{https://orcid.org/0000-0002-5541-437X}{\includegraphics[scale=0.06]{orcid.pdf}\hspace{1mm}Mengwu ~Guo} \\
	Centre for Mathematical Sciences\\
	Lund University, Sweden \\
	\texttt{mengwu.guo@math.lu.se} \\
}

\date{}

\begin{document}
\maketitle

\begin{abstract}
Many physics-informed machine learning methods for PDE-based problems rely on Gaussian processes (GPs) or neural networks (NNs). However, both face limitations when data are scarce, and the dimensionality is high. Although GPs are known for their robust uncertainty quantification in low-dimensional settings, their computational complexity becomes prohibitive as the dimensionality increases. In contrast, while conventional NNs can accommodate high-dimensional input, they often require extensive training data and do not offer uncertainty quantification.
To address these challenges, we propose a PDE-constrained Deep Kernel Learning (PDE-DKL) framework that combines DL and GPs under explicit PDE constraints. Specifically, NNs learn a low-dimensional latent representation of the high-dimensional PDE problem, reducing the complexity of the problem. GPs then perform kernel regression subject to the governing PDEs, ensuring accurate solutions and principled uncertainty quantification, even when available data are limited. This synergy unifies the strengths of both NNs and GPs, yielding high accuracy, robust uncertainty estimates, and computational efficiency for high-dimensional PDEs.
Numerical experiments demonstrate that PDE-DKL achieves high accuracy with reduced data requirements. They highlight its potential as a practical, reliable, and scalable solver for complex PDE-based applications in science and engineering.
\end{abstract}


\section{Introduction}
High-dimensional partial differential equations (PDEs) are indispensable for modelling complex dynamical systems across science and engineering. Unfortunately, traditional numerical methods (e.g., finite element \cite{argyris1979finite}, finite difference \cite{godunov1959finite}, finite volume \cite{eymard2000finite}, and spectral methods \cite{gottlieb1977numerical})
often become computationally intractable in high dimensions and struggle to assimilate sparse data. As a result, there is a growing demand for data-driven techniques that can handle high-dimensional spaces while incorporating observational information efficiently.

Recent advances in deep learning (DL) have shown that NNs can effectively approximate high-dimensional functions \cite{hornik1991approximation}, inspiring methods such as PINNs (physics-informed neural networks) \cite{raissi2019physics, karniadakis2021physics} that add PDE residuals to the loss function to guide the network toward physically consistent solutions.
This explicit residual formulation has further motivated broader strategies to embed physical laws into the learning process \cite{rao2021physics, pang2019fpinns, karpatne2017physics, rao2021hard, zhu2019physics, blechschmidt2021three}.
Despite their promise, most neural-network-based PDE solvers focus on deterministic predictions and do not quantify uncertainty. Many applications, however, demand rigorous uncertainty estimates, especially when observational data are limited. GPs \cite{williams2006gaussian} offer a natural probabilistic framework by providing both predictive mean and variance. Their nonparametric nature helps avoid overfitting with small datasets \cite{mackay1992bayesian}, a key advantage when data acquisition is expensive.

A growing body of work has explored encoding PDE constraints and other priors directly into GP models \cite{swiler2020survey, sarkka2011linear, raissi2017machine, pfortner2022physics}, taking advantage of the fact that Gaussianity is preserved under linear operators \cite{cramer2013stationary}. This has led to kernel-based methods—such as deep kernel learning \cite{wilson2016deep}, manifold GPs \cite{calandra2016manifold}, and guided deep kernel learning \cite{achituve2023guided}—that can learn expressive representations to improve computational scalability and accuracy in higher dimensions.

Combining DNNs with GPs can build hybrid models that offer efficient high-dimensional approximation and principled uncertainty quantification. Such approaches preserve essential physical laws while leveraging data-driven representations, ultimately enabling more accurate and interpretable solutions for high-dimensional PDEs.

\subsection{Related work}
In recent years, progress has been made in the numerical approximation of high-dimensional PDEs using DNNs. Examples include early approaches combining NNs with collocation points \cite{lagaris1998artificial,malek2006numerical}, physics-informed methods that incorporate PDE residuals into the loss function \cite{raissi2019physics, berg2018unified}, and hybrid approaches combining NNs with classical methods, such as finite elements  \cite{ramuhalli2005finite, jung2020deep}.
Moreover, many alternative methods have emerged to address the limitations of traditional approaches. These include the deep Galerkin method \cite{sirignano2018dgm}, which demonstrates effectiveness in high-dimensional problems under regularity conditions, the deep Ritz method \cite{yu2018deep} and its extensions for non-convex domains \cite{liao2019deep}, as well as neural operators \cite{wang2021learning,kovachki2023neural,li2020fourier} that learn mappings between infinite-dimensional function spaces.
While these methods have shown promise in various applications, most struggle to efficiently solve high-dimensional problems due to their inherent complexity and high computational demands \cite{yarotsky2017error}.
The inherent spectral bias of NNs — their tendency to prioritize learning low-frequency functions — fundamentally limits their ability to approximate high-frequency modes critical for resolving sharp gradients in high-dimensional PDEs \cite{wang2021understanding, mishra2023estimates}. A recent study \cite{hao2023pinnacle}, which explores more than ten standard and variant PINN methods, highlights high-dimensional problems that remain challenging for PINNs.

Several efforts to address the computational challenges of high-dimensional systems. For instance, the deep splitting method \cite{beck2021deep} was introduced for the high-dimensional systems, which iteratively solves the linear approximation of the PDE's solution by combining DNN and the Feynman-Kac formula. 
Beck and Berner et al. \cite{beck2018solving, berner2020numerically} also successfully applied the DNN approximation to high-dimensional Kolmogorov PDEs.  
Regarding machine learning approaches to enhance computational efficiency, Nabian et al. \cite{nabian2019deep} enhanced the computational efficiency of randomised NNs by training a deep residual network on residual data. More recently, Wang et al. \cite{wang2024extreme} extended the extreme learning machine approach using an approximate variant of the theory of functional connections (TFC), showing promise in efficiently solving high-dimensional PDEs.
However, despite these advances, most methods still face challenges in efficiently handling complex, high-dimensional problems, especially when ensuring robustness and uncertainty quantification.

Several approaches have been explored to incorporate GP into PDE solvers to address spectral bias while preserving uncertainty quantification.
Many researchers have applied PDE constraints to construct kernels, as exemplified by linearly constrained GPs \cite{jidling2017linearly}, generalised linear solvers \cite{pfortner2022physics}, and multi-fidelity inference \cite{raissi2017inferring}. 
Beckers et al. \cite{beckers2022gaussian} proposed a GP surrogate model using physics priors for port-Hamiltonian systems. Hansen et al. \cite{hansen2023learning} investigated learning physical models respecting conservation laws, and Girolami et al. \cite{girolami2021statistical} synthesised observation data and model predictions using a specific parametric GP prior. 
While effective in handling low-dimensional linear PDEs, these methods often face challenges when extended to high-dimensional nonlinear problems because of the curse of dimensionality and computational demands. The latent force model (LFM) is another important approach that combines GPs with physical principles by modelling the forcing term with a GP prior \cite{sarkka2018gaussian}. However, it still requires solving the underlying differential equation, which limits its scalability. 

Another relevant approach is the neural-net-induced Gaussian process (NNGP) method for PDEs \cite{pang2019neural}. NNGP was introduced to combine DNNs with GP regression for function approximation and solving PDEs. By constructing Mercer kernels through infinite-width DNNs, NNGPs theoretically bridge neural flexibility with GP interpretability. The method achieves NN and GP equivalence by utilising fully connected, infinitely wide DNNs with random weights and biases. This approach employs the NN to induce GP kernels, which require many NN parameters. 
Although NNGP demonstrates advantages over simple GP or DNN models, particularly in flexibility and expressiveness, these benefits have been shown primarily in low-dimensional examples presented in the paper. 

\subsection{Contributions}
Our work has made contributions to the data-driven numerical approximation of high-dimensional PDEs in the following aspects:
\begin{enumerate}
    \item \textbf{Modelling with limited data:} 
    The proposed method effectively models systems with limited observational data, especially for high-dimensional problems. This is particularly useful in real-world applications where data availability is constrained.
    
    \item \textbf{Overcoming the curse of dimensionality:}
    Our research addresses the curse of dimensionality in the numerical approximation of high-dimensional PDEs, a challenge that often leads to an exponential increase in computational costs for traditional methods.
    Such exponential growth is effectively avoided using our method.
    
    \item \textbf{Enhancements in initialisation and optimisation strategies:}
    Our method guarantees computational efficiency by incorporating meticulous initialisation and tailored optimisation strategies. This results in accuracy and improved computational performance.
   
    \item \textbf{Reliability and uncertainty quantification (UQ):}
    Our method harnesses the synergies of DNNs and GPs to enhance accuracy and uncertainty quantification in diverse PDEs. 
    Through a Bayesian framework, the GP component provides built-in uncertainty analysis encoded in the posterior distributions, addressing challenges where modelling uncertainty is often overlooked in conventional DL methods. 
    While the framework is particularly effective for solving linear PDEs, it can also be extended to a wider range of computational tasks.
\end{enumerate}

\subsection{Structure of the paper}
The structure of the paper is as follows: After the introduction, Sections 2 and 3 cover the proposed methodology, including the problem formulation, PDE-constrained GPs, and PDE-constrained DKL. Section 4 presents our model's numerical results, concluding with a discussion of findings. Finally, Section 5 provides concluding remarks by summarising our work and suggesting directions for future research.

\section{Problem formulation and solution methods}
This section introduces the basics of GP regression, the problem formulation considered in this work, and the corresponding GP approximations.

\subsection{Gaussian processes for regression}
In this work, the core idea is to use a GP to describe a distribution over the function space of PDE solutions. In GP regression, the prior of the function to be approximated, denoted by $g: \mathcal{Z} \rightarrow \mathbb{R}$ with $\mathcal{Z}$ being the input domain, is assumed to be a GP, i.e., any finite number of the function outputs follow a joint normal distribution. This GP, written as $g(\cdot)\sim \mathcal{GP}(m(\cdot), k(\cdot, \cdot))$, is uniquely defined by its mean function $m$ and covariance function $k$, respectively given by
\begin{align}
m(\boldsymbol{z}) =\mathbb{E}[g(\boldsymbol{z})]\,, \quad \text{and} \quad 
k\left(\boldsymbol{z}, \boldsymbol{z}^{\prime}\right) =\mathbb{E}\left[(g(\boldsymbol{z})-m(\boldsymbol{z}))\left(g\left(\boldsymbol{z}^{\prime}\right)-m\left(\boldsymbol{z}^{\prime}\right)\right)^\top \right] \,,\quad \vb*{z},\vb*{z}^{\prime}\in \mathcal{Z}\,.
\label{eq:cov}
\end{align}
With this setting, supervised learning is conducted based on a dataset of $N$ input-output pairs $( \vb{Z}, \boldsymbol{y})=\left\{\left(\boldsymbol{z}^{(i)}, y^{(i)}\right)\right\}_{i=1}^N$, in which the observed responses may be corrupted by white noise, i.e., $y^{(i)}=g\left( \boldsymbol{z}^{(i)}\right)+\epsilon_i$ with $\epsilon_i \sim \mathcal{N}\left(0, \sigma_n^2\right)$. 
Thus, the covariance matrix for the vector of observational outputs $\vb*{y}$ is $k(\vb{Z},\vb{Z})+\sigma_n^2 \mathbf{I}$.

Consider any unseen test point $\vb*{z}^* \in \mathcal{Z}$, the joint distribution of the observed and (noise-free) test outputs of $g$ is given by the GP as 
\begin{equation}
\left(\begin{array}{c}
\boldsymbol{y} \\
g(\vb*{z}^*)
\end{array}\right) \sim \mathcal{N}\left(\left(\begin{array}{l}
m(\vb{Z}) \\
m(\vb*{z}^*)
\end{array}\right),\left(\begin{array}{cc}
k(\vb{Z},\vb{Z})+\sigma_n^2 \mathbf{I}  & k(\vb{Z},\vb*{z}^*) \\
k(\vb*{z}^*,\vb{Z}) & k(\vb*{z}^*,\vb*{z}^*)
\end{array}\right)\right)\,.
\label{eq:joint_density}
\end{equation}
To update the prior GP to agree with observations, the rule of conditional Gaussian gives that the predictive distribution of $g(\vb*{z}^*)$ is Gaussian again, i.e., $g(\vb*{z}^*) \mid ( \vb{Z}, \boldsymbol{y}) \sim \mathcal{N}\left(e^*, v^*\right)$, in which 
\begin{equation} \label{eq:postgp}
\begin{split}
e^*
&= m\left(\vb*{z}^*\right) + k\left(\vb*{z}^* , \vb{Z}\right)\left(k(\vb{Z},\vb{Z})+\sigma_n^2 \mathbf{I}\right)^{-1} \left(\boldsymbol{y} - m(\vb{Z}) \right)\,,\quad \text{and} \\
v^* &= k\left(\vb*{z}^*, \vb*{z}^*\right) - k\left(\vb*{z}^*, \vb{Z}\right)\left(k(\vb{Z},\vb{Z})+\sigma_n^2 \mathbf{I}\right)^{-1}k\left(\vb{Z}, \vb*{z}^* \right)\,.
    \end{split}
\end{equation}
One significant advantage of GP regression is that the probabilistic predictive model has analytical formulations.

\subsection{Model problem for high-dimensional PDEs}
We consider the spatial-temporal-parametric coordinates $\vb*{q}=(\vb*{x}, t, \boldsymbol{\mu})\in \Omega\times \mathcal{T}\times\mathcal{M}:=\mathcal{Q}$, with $\Omega\subset \mathbb{R}^{n_x}$, $\mathcal{T}\subset \mathbb{R}$, and $\mathcal{M}\subset \mathbb{R}^{n_\mu}$ respectively denoting the space, time, and parameter domains, and $\mathcal{Q}\subset \mathbb{R}^{n_q}$ their joint domain in  $n_q = n_x +1+n_\mu$ dimension. The high-dimensional forward problems considered in this work are governed by spatial-temporal-parametric PDEs generally written in the following form: 
\begin{equation}
   \partial_t u(\vb*{q})=\mathcal{L}_{\vb*{x}}[u(\vb*{q}); \vb*{\mu}]+f(\vb*{q}), \quad\text{with}\quad
   \mathcal{L}_{\vb*{x}}[\cdot; \vb*{\mu}]:=\sum_{\alpha \in \mathbb{N}} C_\alpha(\vb*{x};\vb*{\mu}) \frac{\partial^\alpha}{\partial \vb*{x}^\alpha}\,,
   \label{eq:pde}
\end{equation}
equipped with proper boundary and initial conditions.
Specifically, $u: \mathcal{Q} \rightarrow \mathbb{R}$ is the solution field, and $\mathcal{L}_{\vb*{x}}[\cdot ; \vb*{\mu}]$ is a linear differential operator over space parametrized by $\vb*{\mu}$. 
We combine $\mathcal{L}_{\vb*{x}}$ and the time derivative $\partial_t$, both being linear operators acting on the solution field $u$, into a spatial-temporal parametric linear operator $ \mathcal{A} [u]:=\partial_t u-\mathcal{L}_{\vb*{x}}[u ; \boldsymbol{\mu}]$. Therefore, the governing equation \eqref{eq:pde} is rewritten as follows:
\begin{equation}
    \mathcal{A} [u({\vb*{q}})]=f( {\vb*{q}}),
    \label{eq:simplify_pde}
\end{equation}
where $f: \mathcal{Q} \rightarrow \mathbb{R}$ is a given function for a forcing term. 

\subsection{Constraining Gaussian processes with the PDEs model} \label{constrain_gp}
We adopt the aforementioned GP regression model as a surrogate for the latent solution function $u$. In this subsection, based on the fact that Gaussianity is preserved under linear operations, we discuss constraining GP emulation with the model problem's governing PDE.

Assume that the latent solution $u$ follows a (prior) GP over the domain of interest $\mathcal{Q}$: 
\begin{equation}
u({\vb*{q}}) \sim \mathcal{G P}^{[u]}\left(m({\vb*{q}}), k\left({\vb*{q}}, {\vb*{q}}^{\prime} \right)\right) \,, \quad \vb*{q},\vb*{q}'\in \mathcal{Q}\,,
    \label{eq:u_gp}
\end{equation}
in which $m:\mathcal{Q}\to \mathbb{R}$ and $k:\mathcal{Q} \times \mathcal{Q}\to \mathbb{R}$ are respectively the mean and covariance functions. According to the governing PDE \eqref{eq:simplify_pde}, the forcing term $f$, as the result of a linear operation $\mathcal{A}$ acting on the latent GP $u$, is also a GP over $\mathcal{Q}$,
it has a GP prior, which is modified according to the operator:
\begin{equation}
    f({\vb*{q}}) = \mathcal{A}[u({\vb*{q}})] \sim \mathcal{G} \mathcal{P}^{[f]}\left(m_{\mathcal{A}}({\vb*{q}}), k_{\mathcal{A}}\left({\vb*{q}}, {\vb*{q}}^{\prime} \right)\right)\,, \quad \vb*{q},\vb*{q}'\in \mathcal{Q}\,,
    \label{eq:f_gp}
\end{equation}
whose mean $m_{\mathcal{A}}$ and covariance $k_{\mathcal{A}}$ are correspondingly defined by applying the operator $\mathcal{A}$ to $m$ and $k$ in $\mathcal{G P}^{[u]}$, i.e.,
\begin{equation}
    \begin{aligned}
    m_{\mathcal{A}}({\vb*{q}}) & = \mathbb{E}[\mathcal{A}[u({\vb*{q}})]]=\mathcal{A} m({\vb*{q}})\,, \\
    \vspace{5mm}
    k_{\mathcal{A}}\left({\vb*{q}}, {\vb*{q}}^{\prime} \right) & =\mathbb{E}\left[(\mathcal{A}[u({\vb*{q}})] -  m_{\mathcal{A}}({\vb*{q}}) )(\mathcal{A}\left[u\left({\vb*{q}}^{\prime}\right)\right] -  m_{\mathcal{A}}({\vb*{q}^{\prime}}))^\top\right] 
    =\mathcal{A}  k\left({\vb*{q}}, {\vb*{q}}^{\prime}\right)\mathcal{A}^\top\,.
    \end{aligned}
    \label{eq:convert}
\end{equation}
Here, the symbol $\mathcal{A}^\top$ represents that the differential operator $\mathcal{A}$ is acting on the kernel function $k$ concerning the second argument. This way, the PDE constraint is explicitly incorporated into the GP model.

In the model problem setting \eqref{eq:pde}, the forcing term $f({\vb*{q}})$ is a given function, from which data at different $\vb*{q}$-locations can be directly generated, and additional measurements for the latent solution $u$ can be evaluated from the boundary and initial conditions. In practice, we may also attain several possibly noise-corrupted observations of $u$ inside the domain $\mathcal{Q}$. With all these available data, we can approximate the entire solution field $u$ via GP regression in a supervised learning manner, as the physical information has been encompassed through the linear transformation from $u$ to $f$. 
Hence, we let $({\vb{Q}}_u, \mathbf{y}_u)= \{({{\vb*{q}}_u^{(i)}, {y}_u^{(i)})}\}_{i=1}^N$ denote the input-output pairs for the observations of $u$, including both the data evaluated from the boundary/initial conditions and those collected inside the domain, let $({\vb{Q}}_f, \mathbf{y}_f )= \{({{\vb*{q}}_f^{(j)}, {y}_f^{(j)} })\}_{j=1}^M$ represent the data pairs computed from the given function $f$, and let $\mathcal{D}_{uf}$ denote the combined dataset.
Hence, the aforementioned correlation between the solution $u$ and force $f$, both being GPs in this case, gives the following joint normal distribution for the observations:
\begin{equation}
   \left[\begin{array}{c}
    \mathbf{y}_u \\
    \mathbf{y}_f
    \end{array}\right] \sim \mathcal{N}\left(\left[\begin{array}{r}
    m\left({\vb{Q}}_{u}\right) \\
    \mathcal{A} m\left({\vb{Q}}_f\right)
    \end{array}\right], {\underbrace {\left[\begin{array}{rr}
    k\left({\vb{Q}}_u, {\vb{Q}}_u\right)+{\sigma_u^2 \vb{I}} &  k\left({\vb{Q}}_u, {\vb{Q}}_f\right)\mathcal{A}^\top \\
    \mathcal{A} k\left({\vb{Q}}_f, {\vb{Q}}_u\right) & \mathcal{A}  k\left({\vb{Q}}_f, {\vb{Q}}_f\right)\mathcal{A}^\top+{\sigma_f^2 \vb{I}}
    \end{array}\right]}_{:=\mathbf{K}}}\right) \,,
    \label{eq:joint}
\end{equation}
in which $\mathbf{K}$ denotes the covariance matrix. Here, we add white noise terms with variances $\sigma_u^2$ and $\sigma_f^2$ to the observations of $u$ and $f$, respectively, to model potential corruptions in the data and/or to improve the condition number of $\vb{K}$.

Conditioning on the observed data in $\mathcal{D}_{uf}$, we can proceed to obtain probabilistic predictions for the solution values of $u$ at any spatial-temporal-parametric location $\vb*{q}\in \mathcal{Q}$; $u^*(\vb*{q})\mid \mathcal{D}_{uf} \sim \mathcal{GP} \left( m^*({\vb*{q}}), k^{*}({\vb*{q}},\vb*{q}')\right)$. We explicitly write the mean and covariance functions for this predictive GP as follows:
\begin{equation}
    \begin{aligned}
    m^*({\vb*{q}}) & = m\left({\vb*{q}}\right) + 
    \left[
    k\left({\vb*{q}}, {\vb{Q}}_u\right) \hspace{0.5mm} 
    k\left({\vb*{q}}, {\vb{Q}}_f\right)\mathcal{A}^\top
    \right]
    \mathbf{K}^{-1} 
\left[\begin{array}{l}
\mathbf{y}_u-m\left(
\vb{Q}_{u}\right) \\
\mathbf{y}_f-\mathcal{A}m\left(\vb{Q}_{f}\right)
\end{array}\right]
\,,\\
   k^*({\vb*{q}},\vb*{q}') & = k\left({{\vb*{q}}}, {{\vb*{q}}}'\right)- \left[k\left({\vb*{q}}, {\vb{Q}}_u\right) \hspace{0.5mm} k\left({\vb*{q}}, {\vb{Q}}_f\right)\mathcal{A}^\top\right]\mathbf{K}^{-1}\left[k\left({\vb*{q}}', {\vb{Q}}_u\right) \hspace{0.5mm} k\left({\vb*{q}}', {\vb{Q}}_f\right)\mathcal{A}^\top\right]^{\top} \,.
   \end{aligned}
   \label{eq:joint_new_u}
\end{equation}
In fact, one can also derive the reconstruction of $f$ as a GP, i.e., $f^*(\vb*{q}) \mid \mathcal{D}_{uf}\sim \mathcal{GP}(m_{\mathcal{A}}^*({\vb*{q}}),k_{\mathcal{A}}^*({\vb*{q}},\vb*{q}'))$, in which
\begin{equation}
    \begin{aligned}
    m_{\mathcal{A}}^*({\vb*{q}}) & = {\mathcal{A}}m
    \left({\vb*{q}}\right) + 
    \left[
     \mathcal{A}k\left({\vb*{q}}, {\vb{Q}}_u\right) \hspace{0.5mm} \mathcal{A}
    k\left({\vb*{q}}, {\vb{Q}}_f\right)\mathcal{A}^\top
    \right]
    \mathbf{K}^{-1} \left[\begin{array}{l}
\mathbf{y}_u-m\left(
\vb{Q}_{u}\right) \\
\mathbf{y}_f-\mathcal{A}m\left(\vb{Q}_{f}\right)
\end{array}\right] \quad \,,\\
   k_{\mathcal{A}}^*({\vb*{q}},\vb*{q}') & = {\mathcal{A}}k\left({{\vb*{q}}}, {{\vb*{q}}}'\right){\mathcal{A}}^\top - \left[\mathcal{A}k\left({\vb*{q}}, {\vb{Q}}_u\right) \hspace{0.5mm} \mathcal{A}k\left({\vb*{q}}, {\vb{Q}}_f\right)\mathcal{A}^\top\right]\mathbf{K}^{-1}\left[\mathcal{A}k\left({\vb*{q}}', {\vb{Q}}_u\right) \hspace{0.5mm} \mathcal{A}k\left({\vb*{q}}', {\vb{Q}}_f\right)\mathcal{A}^\top\right]^{\top}\,.
   \end{aligned}
   \label{eq:joint_new_f}
\end{equation}
It is evident that 
\begin{equation}
    m_{\mathcal{A}}^*({\vb*{q}}) = \mathcal{A}m^*({\vb*{q}})\,,\quad \text{and}\quad
    k_{\mathcal{A}}^*({\vb*{q}},\vb*{q}') = \mathcal{A}k^*({\vb*{q}},\vb*{q}') \mathcal{A}^\top\,,
    \label{eq:relation}
\end{equation}
implying that the pair of solved $u$ and reconstructed $f$, both being GPs, satisfies the constraint of linear transformation \eqref{eq:simplify_pde} through $\mathcal{A}$. Such a constraint is preserved throughout the GP surrogate modelling from the prior setting \eqref{eq:f_gp} to the posterior prediction \eqref{eq:relation}, confirming the PDE-constrained nature of this approximation method. 

\section{Deep kernel learning with PDE constraints}

\subsection{Construction of a deep kernel}\label{sec:DK}
The choice of kernel in GP regression is crucial as it directly affects the performance of function approximation. In this work, to improve the kernel's expressive power in high-dimensional input space, we use a manifold GP \cite{calandra2016manifold} in which a multilayer perceptron NN is embedded into a conventionally used kernel. 
In particular, we consider the following squared exponential kernel with automatic relevance determination (ARD) as the `base kernel':
\begin{equation}
    {k}_\texttt{ARD}\left(\vb*{h}, {\vb*{h}}'; \boldsymbol{\theta}\right) =\sigma^2 \exp \left(-\frac{1}{2} ({\vb*{h}}-{\vb*{h}'})^{\top} \text{diag}(\vb*{\ell}^2)^{-1} ({\vb*{h}}-{\vb*{h}'})\right) \,,
\end{equation}
defined in $\mathbb{R}^{n_h}$, i.e., $\vb*{h},\vb*{h}'\in \mathbb{R}^{n_h}$. 
The hyperparameters $\boldsymbol{\theta}=\left\{\sigma^2, \vb*{\ell}\right\}$ in this kernel include a global variance $\sigma^2$ and individual lengthscales collected in $\vb*{\ell}=\left\{\ell_1, \ell_2, \cdots, \ell_{n_h}\right\}^{\top}$, each capturing the variability along a specific dimension.
Note that this ARD kernel is smooth, anisotropic, and infinitely differentiable, and the so-called Mahalanobis distance \cite{mahalanobis2018generalized} is measured. 

\begin{figure}  \includegraphics[width=\linewidth,height=\textheight,keepaspectratio]{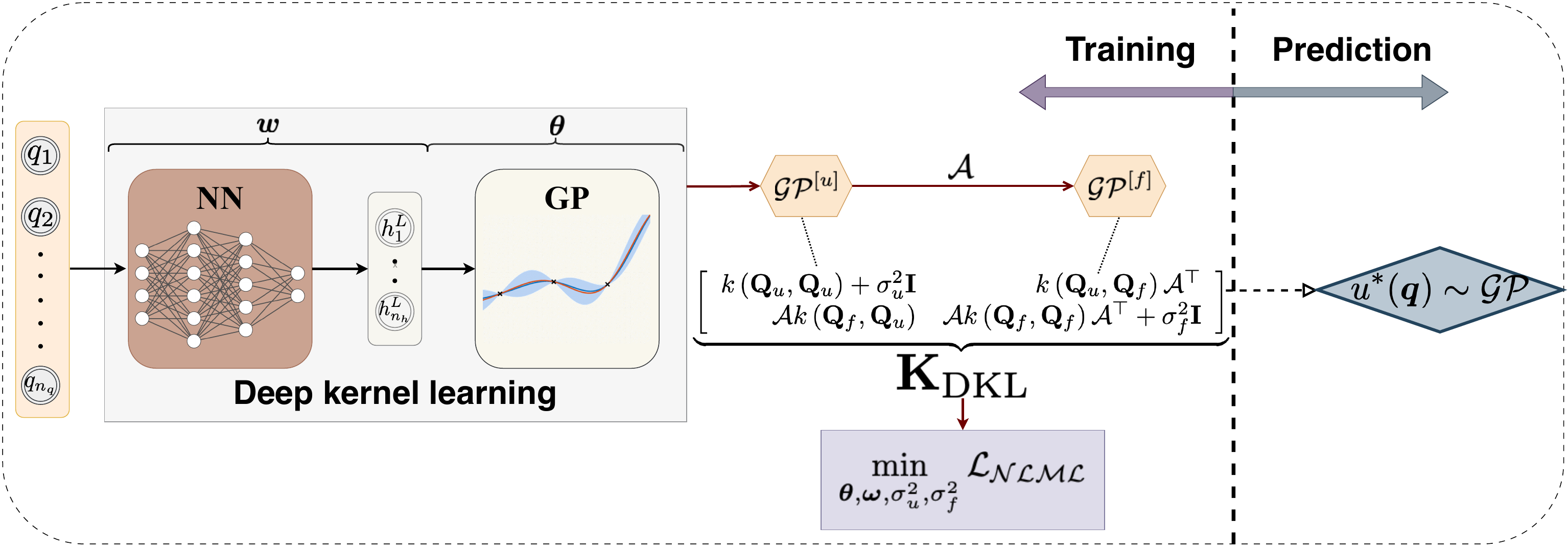}
    \caption{Conceptual diagram of PDE-constrained deep kernel learning (PDE-DKL): The initialisation training involves using PDE coordinates $\vb*{q}$ as inputs to a deep kernel. This kernel is formed by integrating an NN with parameters $\vb*{\omega}$ into a base kernel with hyperparameters $\vb*{\theta}$. It serves as the covariance function for the latent GP solution $u$. Applying a linear operator $\mathcal{A}$ to the latent GP produces a GP for the forcing term $f$. Subsequently, the joint kernel $\mathbf{K}_{\texttt{DKL}}$ is employed to minimize the negative log marginal likelihood $\mathcal{L}_{\mathcal{N L M L}}$, determining the parameters $\left\{\vb*{\theta}, \vb*{\omega}, \sigma_u^2, \sigma_f^2\right\}$. The trained model evaluates the solution field $u^*(\boldsymbol{q})$ in the prediction stage.}
\label{fig:dkl_slides-Page-21}
\end{figure}

We further compose this base kernel with an $L$-layer NN $\vb*{h}^L(\cdot;\vb*{\omega}): \mathcal{Q} \to \mathbb{R}^{n_h}$, whose inputs are the PDE coordinates $\vb*{q}$, and the last layer has $n_h$ units (Figure~\ref{fig:dkl_slides-Page-21}). Here, $\vb*{\omega}$ denotes a vector that collects the NN's parameters (i.e., weights and biases). Therefore, the composed kernel often referred to as a `deep kernel' \cite{wilson2016deep}, can be written as
\begin{equation}
        {k}_\texttt{DKL}\left({\vb*{q}}, {\vb*{q}}^{\prime}; \boldsymbol{\theta}, \boldsymbol{\omega}\right) =\sigma^2 \exp \left(-\frac{1}{2} (\boldsymbol{h}^L(\vb*{q};\boldsymbol{\omega})-\boldsymbol{h}^L\left(\vb*{q}^{\prime};\boldsymbol{\omega}\right))^{\top} \text{diag}(\vb*{\ell}^2)^{-1} (\boldsymbol{h}^L(\vb*{q};\boldsymbol{\omega})-\boldsymbol{h}^L\left(\vb*{q}^{\prime};\boldsymbol{\omega}\right))\right) \,,
    \label{eq:ard}
\end{equation}
by taking $\boldsymbol{h}=\boldsymbol{h}^L(\vb*{q};\boldsymbol{\omega})$ and $\boldsymbol{h}'=\boldsymbol{h}^L\left(\vb*{q}^{\prime};\boldsymbol{\omega}\right)$, i.e., the NN outputs at the input values $\vb*{q}$ and $\vb*{q}'$, respectively.

By incorporating the NN, we upgraded a stationary kernel to a non-stationary one, significantly improving nonlinear expressiveness and scalability to high-dimensional function approximation. On the other hand, the subsequent `GP layer' facilitates NN modelling with uncertainty quantification. 

High-dimensional function approximation is known to be challenging for GP regression, as the latter often suffers from the curse of dimensionality when conventional kernels are used. A strategy to overcome this is to find a faithful data representation in a low-dimensional latent space. Our treatment is aligned with this spirit. When the input domain $\mathcal{Q}$ has a high dimensionality $n_q$, the use of a `deep kernel' allows us to transform the size-$n_q$ input data into a $n_h$-dimensional latent space through a NN; in fact, this is a nonlinear dimensionality reduction of the inputs, or referred to as a \emph{sensitivity analysis}. Following this, a GP regression with the ARD kernel is conducted in the $n_h$-dimensional latent space. When $n_h$ is sufficiently small, the GP regression will have guaranteed performance and does not suffer from the curse of dimensionality.

\subsection{Constraining deep kernel with PDEs}
When the aforementioned `deep kernel' is used to approximate the solution field $u$ of a high-dimensional PDE problem \eqref{eq:simplify_pde}, (i.e., $n_q \gg 1$), the regression through deep kernel learning in subsection \ref{sec:DK} needs to be constrained by \eqref{eq:simplify_pde}, so that information from the governing physics is integrated into the DL model. As the deep kernel learning method still falls under the realm of GP regression, this allows us to integrate PDE constraints \eqref{eq:simplify_pde} into the deep kernel \eqref{eq:ard}, treated similarly with subsection \ref{constrain_gp}. 

In this case, the latent solution $u$ is assumed to follow a prior GP induced by a deep kernel, i.e., $u_\texttt{DKL}(\vb*{q})\sim \mathcal{GP}(m_\texttt{DKL}(\vb*{q}),k_\texttt{DKL}(\vb*{q},\vb*{q}'))$, where $m_\texttt{DKL}(\vb*{q})$ represents the mean function of the GP and is typically set to zero (or another appropriate prior mean if available). The covariance function $k_\texttt{DKL}(\vb*{q},\vb*{q}')$ is defined by the deep kernel, which is constructed through a NN architecture, effectively learning a data-driven similarity measure between inputs $\vb*{q},\vb*{q}' $. Similar to \eqref{eq:convert}, we can reconstruct the forcing term $f$ by applying a linear operation $\mathcal{A}$ on the latent GP $u_\texttt{DKL}$. Therefore,  the forcing term $f$ can be reconstructed as
$
    f_\texttt{DKL}({\vb*{q}}) = \mathcal{A}[u_\texttt{DKL}({\vb*{q}})] \sim \mathcal{G} \mathcal{P}^{[f]}\left(\mathcal{A}m_{{{\texttt{DKL}}}}({\vb*{q}}), \mathcal{A}k_{{{\texttt{DKL}}}}\left({\vb*{q}}, {\vb*{q}}^{\prime} \right)\mathcal{A}^\top\right) 
    \label{eq:f_dkl}
$. 

Directly using the results described in subsection (\ref{constrain_gp}), the predictive mean and covariance functions for the DKL solution field $u_\texttt{DKL}^*(\vb*{q})\mid \mathcal{D}_{uf} \sim \mathcal{GP}(m_{\texttt{DKL}}^*({\vb*{q}}),k_{\texttt{DKL}}^*({\vb*{q}},\vb*{q}'))$ are respectively given by
\begin{equation}
    \begin{aligned}
    m_\texttt{DKL}^*({\vb*{q}}) & = m_\texttt{DKL}\left({\vb*{q}}\right) + 
    \left[
    k_\texttt{DKL}\left({\vb*{q}}, {\vb{Q}}_u\right) \hspace{0.5mm} 
    k_\texttt{DKL}\left({\vb*{q}}, {\vb{Q}}_f\right)\mathcal{A}^\top
    \right]
    \mathbf{K}_\texttt{DKL}^{-1} \left[\begin{array}{l}
\mathbf{y}_u-m_\texttt{DKL}\left(
\vb{Q}_{u}\right) \\
\mathbf{y}_f-\mathcal{A}m_\texttt{DKL}\left(\vb{Q}_{f}\right)
\end{array}\right] \quad \text{and}\\
   k_\texttt{DKL}^*({\vb*{q}},\vb*{q}') & = k_\texttt{DKL}\left({{\vb*{q}}}, {{\vb*{q}}}'\right)- \left[k_\texttt{DKL}\left({\vb*{q}}, {\vb{Q}}_u\right) \hspace{0.5mm} k_\texttt{DKL}\left({\vb*{q}}, {\vb{Q}}_f\right)\mathcal{A}^\top\right]\mathbf{K}_\texttt{DKL}^{-1}\left[k_\texttt{DKL}\left({\vb*{q}}', {\vb{Q}}_u\right) \hspace{0.5mm} k_\texttt{DKL}\left({\vb*{q}}', {\vb{Q}}_f\right)\mathcal{A}^\top\right]^{\top}, \\
   \text{with} \quad \mathbf{K}_\texttt{DKL} &= \left[\begin{array}{rr}
    k_\texttt{DKL}\left({\vb{Q}}_u, {\vb{Q}}_u\right)+{\sigma_u^2 \vb{I}} &  k_\texttt{DKL}\left({\vb{Q}}_u, {\vb{Q}}_f\right)\mathcal{A}^\top \\
    \mathcal{A} k_\texttt{DKL}\left({\vb{Q}}_f, {\vb{Q}}_u\right) & \mathcal{A}  k_\texttt{DKL}\left({\vb{Q}}_f, {\vb{Q}}_f\right)\mathcal{A}^\top+{\sigma_f^2 \vb{I}}
    \end{array}\right] \,,
   \end{aligned}
   \label{eq:joint_new_u_dkl}
\end{equation}
Here, the kernel $k$ in \eqref{eq:joint_new_u} is specified to be the deep kernel $k_\texttt{DKL}$ given by \eqref{eq:ard}.
Then we can reconstruct $f_{\mathcal{A}({\texttt{DKL}})}^*(\vb*{q}) \mid \mathcal{D}_{uf} \sim \mathcal{GP}(m_{\mathcal{A}({\texttt{DKL}})}^*({\vb*{q}}),k_{\mathcal{A}({\texttt{DKL}})}^*({\vb*{q}},\vb*{q}'))$ with
\begin{equation}
    \begin{aligned}
    m_{\mathcal{A}({\texttt{DKL}})}^*({\vb*{q}}) & = m_{\mathcal{A}({\texttt{DKL}})} \left({\vb*{q}}\right) + 
    \left[
     \mathcal{A}k_\texttt{DKL}\left({\vb*{q}}, {\vb{Q}}_u\right) \hspace{0.5mm} \mathcal{A}
    k_\texttt{DKL}\left({\vb*{q}}, {\vb{Q}}_f\right)\mathcal{A}^\top
    \right]
    \mathbf{K}_\texttt{DKL}^{-1} \left[\begin{array}{l}
\mathbf{y}_u-m_\texttt{DKL}\left(
\vb{Q}_{u}\right) \\
\mathbf{y}_f-\mathcal{A}m_\texttt{DKL}\left(\vb{Q}_{f}\right)
\end{array}\right] \quad \,,\\
   k_{\mathcal{A}({\texttt{DKL}})}^*({\vb*{q}},\vb*{q}') & = k_{\mathcal{A}({\texttt{DKL}})}\left({{\vb*{q}}}, {{\vb*{q}}}'\right)- \left[\mathcal{A}k_\texttt{DKL}\left({\vb*{q}}, {\vb{Q}}_u\right) \hspace{0.5mm} \mathcal{A}k_\texttt{DKL}\left({\vb*{q}}, {\vb{Q}}_f\right)\mathcal{A}^\top\right]\mathbf{K}_\texttt{DKL}^{-1}\left[\mathcal{A}k_\texttt{DKL}\left({\vb*{q}}', {\vb{Q}}_u\right) \hspace{0.5mm} \mathcal{A}k_\texttt{DKL}\left({\vb*{q}}', {\vb{Q}}_f\right)\mathcal{A}^\top\right]^{\top} \,,
   \end{aligned}
   \label{eq:joint_new_f_dkl}
\end{equation}
and the relations in \eqref{eq:relation} still hold.

This treatment retains the nature of GP surrogate modelling, enabling epistemic uncertainty quantification for solving high-dimensional PDEs with deep kernel functions. By incorporating the PDE constraints, we leverage prior physical knowledge and ensure the physical interpretability of this probabilistic DL method. Moreover, the model training also benefits significantly from the effective use of the data from both observations ($u$-data) and PDEs ($f$-data).

Given a deep kernel, both the assembly of $\vb{K}_\texttt{DKL}$ and the evaluation of \eqref{eq:joint_new_u_dkl} require derivative evaluations of the kernel function $k_\texttt{DKL}$ concerning the inputs $\vb*{q}$ when the linear differential operator $\mathcal{A}$ is applied.
The derivatives can be computed using the chain rule, i.e.,
\begin{equation}
    \frac{\partial k_\texttt{DKL}}{\partial \vb*{q}}=\frac{\partial k_\texttt{DKL}}{\partial \boldsymbol{h}^L} \cdot \frac{\partial \boldsymbol{h}^L}{\partial \vb*{q}}\,.
    \label{derivative}
\end{equation}
Here, $\frac{\partial k_{\mathrm{DKL}}}{\partial \boldsymbol{h}^L}$ represents the gradient of the kernel function with respect to the latent variables $h^L$, which is a row vector of size $1 \times n_h$, where $n_h$ is the dimensionality of the latent space. $\frac{\partial \boldsymbol{h}^L}{\partial \boldsymbol{q}}$ is the Jacobian matrix of the NN output $\boldsymbol{h}^L$ with respect to the input $\boldsymbol{q}$, and its size is $n_h \times n_q$, where $n_q$ is the number of features of the input $\boldsymbol{q}$.
Applying the chain rule, by multiplying the row vector $\frac{\partial k_{\mathrm{DKL}}}{\partial \boldsymbol{h}^L}$ with the Jacobian $\frac{\partial \boldsymbol{h}^L}{\partial \boldsymbol{q}}$, results in the gradient $\frac{\partial k_{\mathrm{DKL}}}{\partial \boldsymbol{q}}$, which is a vector of size $1 \times n_q$. This gradient captures how the kernel function changes for each component of the input $\boldsymbol{q}$. Practically, $\frac{\partial k_{\mathrm{DKL}}}{\partial \boldsymbol{h}^L}$ can be written analytically from the form of the base ARD kernel, and $\frac{\partial \boldsymbol{h}^L}{\partial \boldsymbol{q}}$ is efficiently evaluated by (backward) automatic differentiation through the NN.

Handling high-dimensional inputs $\boldsymbol{q}$ presents several fundamental challenges. First, as the dimensionality of $\boldsymbol{q}$ increases, the computational complexity associated with calculating the Jacobian $\frac{\partial \boldsymbol{h}^L}{\partial \boldsymbol{q}}$ grows rapidly, given that each dimension introduces additional partial derivatives that must be evaluated. This scaling issue becomes particularly acute when the latent space $\boldsymbol{h}^L$ is also large, resulting in significant computational overhead. Additionally, this computation is relatively straightforward when dealing with first-order derivatives. However, the complexity significantly increases if the linear operator $\mathcal{A}$ involves higher-order derivatives. For instance, the evaluation of
$
\mathcal{A} k_{\texttt{DKL}}(\boldsymbol{q}, \cdot) \mathcal{A}^{\top}
$
requires calculating second-order spatial derivatives of the kernel concerning both arguments. This will greatly escalate the computational demands, including storage and run time. Second, storing these large Jacobian matrices for each data point leads to considerable memory demands. Moreover, high-dimensional spaces are susceptible to numerical instability, where gradient-based methods may encounter issues such as vanishing or exploding gradients, adversely affecting both the convergence and accuracy of the model. Finally, the curse of dimensionality exacerbates these problems, as high-dimensional inputs often lead to sparse data distributions, making it increasingly difficult for the model to generalise effectively across all dimensions. These combined challenges necessitate sophisticated optimisation techniques and careful architectural design to mitigate computational and numerical difficulties when dealing with high-dimensional data.

To mitigate these challenges, we introduce specialised algorithms that directly address the complexities of higher-order derivative evaluations in high-dimensional spaces. Specifically, these algorithms exploit the sparsity in the Jacobian and Hessian structures and utilise techniques such as automatic differentiation and parallelisation to reduce both the computational cost and memory footprint. In the context of the DKL framework, these optimisations are carefully designed to maintain the accuracy of kernel evaluations while handling large-scale, high-dimensional data. The following sections will detail the development and implementation of these modelling steps, providing specific examples of how they improve computational efficiency in different cases.

\subsection{Model training}
The proposed DKL model includes several trainable parameters, denoted as $\{\boldsymbol{\theta}, \boldsymbol{\omega}\}$. In our approach, effective model training should not only be robust to high-dimensional data but also ensure strong generalisation performance. Our goal is to leverage both data pairs, $\left(\mathbf{Q}_u, \mathbf{y}_u\right)$ and $\left(\mathbf{Q}_f, \mathbf{y}_f\right)$, to jointly perform inference and prediction of the solution. Specifically, we first use Bayesian optimisation to determine the optimal dimensionality of the latent space $\boldsymbol{h}^L$. Following this, we apply type-II marginal likelihood \eqref{eq:lnlml} as the loss function to jointly learn all model parameters $\{\boldsymbol{\theta}, \boldsymbol{\omega}\}$.
\paragraph{Bayesian optimisation}\label{BayesOp}
In this work, we expect to extract a low latent dimensionality from the data for the model inputs $\boldsymbol{q}$, i.e., a nonlinear sensitivity analysis can effectively reduce the input dimensionality. Hence, in our DKL model, the reduced inputs for GP emulation are the values $\boldsymbol{h}^L(\boldsymbol{q})$ at the last network layer. A smaller size of $\boldsymbol{h}^L$ can save the GP with the base kernel from the curse of dimensionality.
To determine the optimal latent dimensions, we tune the hyperparameters for the size of the last NN layer. Given the high computational cost of multiple training iterations, traditional methods like random or grid search are impractical. 
Therefore, we employ Bayesian optimisation to search for the optimal value of $n_h^L$ efficiently. The detailed procedure is outlined in (Algorithm \ref{alg:bayesop}).
Bayesian optimisation is highly efficient for hyperparameter tuning because it can locate an optimum with fewer function evaluations. Our approach employs the negative log marginal likelihood as the objective function, allowing us to optimise the model without needing a separate validation set or objective function. 
By defining an objective function that depends on $n_h^L$ and aims to minimise the $\mathcal{L}_{\mathcal{N L M} \mathcal{L}}$, we iteratively evaluate different values of $n_h^L$ and update the GP surrogate model based on the observed $\mathcal{L}_{\mathcal{N L M} \mathcal{L}}$ values. This approach effectively navigates the hyperparameter space to identify the global optimum. Integrating Bayesian optimisation into the training process of our PDE-constrained DKL model enhances both efficiency and model performance by reducing computational costs and improving the likelihood of finding the optimal hyperparameter configuration.

\paragraph{Loss function}
The DKL model is a GP; therefore, we use the negative log marginal likelihood (NLML), also known as empirical Bayes, to jointly determine the kernel hyperparameters $\boldsymbol{\theta}$ and the NN parameters $\boldsymbol{\omega}$. Initially, we employ Bayesian optimisation to discover potential latent dimensions, which serve as the starting point for learning the network parameters. The kernel hyperparameters are initialised using (Algorithm \ref{alg:init}). We then optimize $\{\boldsymbol{\omega}, \boldsymbol{\theta}, \sigma_u^2, \sigma_f^2\}$ by minimizing the NLML \eqref{eq:lnlml} using gradient descent, where $\sigma_u^2$ and $\sigma_f^2$ represent the noise variances associated with the observations $\mathbf{y}_u$ and $\mathbf{y}_{f}$ respectively.
With the mean vector defined as 
$\vb{m}:= [m\left({\vb{Q}}_u\right)\mathcal{A} m\left({\vb{Q}}_f\right)]^\top$, the objective function is written as 
\begin{equation}
    \begin{split}
    &~ {\mathcal{L}_{\mathcal{N L M L}}(\vb*{\theta},\vb*{\omega},\sigma_u^2,\sigma_f^2)} =  
   -\log p\left(\mathbf{y}_u, \mathbf{y}_f \mid {\vb{Q}}_u, {\vb{Q}}_f, \boldsymbol{\theta},\boldsymbol{\omega}, \sigma_u^2,\sigma_f^2\right) \\ 
    = &~ \frac{1}{2}(\mathbf{y}_{uf}-\mathbf{m})^{\top} \mathbf{K}_\texttt{DKL}(\vb*{\theta},\vb*{\omega},\sigma_u^2,\sigma_f^2)^{-1}(\mathbf{y}_{uf}-\mathbf{m})+\frac{1}{2} \log \left|\mathbf{K}_\texttt{DKL}(\vb*{\theta},\vb*{\omega},\sigma_u^2,\sigma_f^2,)\right|+\frac{N+M}{2} \log (2 \pi) \,,
    \end{split}
    \label{eq:lnlml}
\end{equation} 
where $\mathbf{y}_{uf}=\binom{\mathbf{y}_u}{\mathbf{y}_f}$ is the concatenated vector of observed outputs, and $\mathbf{K}_{\texttt{DKL}}$ is the kernel matrix defined explicitly as \eqref{eq:joint_new_u_dkl}.
The detailed procedure for optimising the trainable parameters with PDE constraints is outlined in
(Algorithm \ref{alg: optimisation}). This algorithm describes the iterative process of minimising the NLML using gradient descent, updating the parameters $\boldsymbol{\omega}, \boldsymbol{\theta}, \sigma_u^2$ and $\sigma_f^2$ until convergence.

In gradient descent, we use the chain rule to compute derivatives to the parameters as follows:
\begin{equation}
        \frac{\partial \mathcal{L}_\mathcal{NLML}}{\partial \boldsymbol{\omega}} =\frac{\partial \mathcal{L}_\mathcal{NLML}}{\partial  \mathbf{K}_\texttt{DKL}} \frac{\partial \mathbf{K}_\texttt{DKL}}{\partial \boldsymbol{h}^L} \frac{\partial \boldsymbol{h}^L}{\partial \boldsymbol{\omega}}\,,\quad 
        \frac{\partial \mathcal{L}_\mathcal{NLML}}{\partial\{ \boldsymbol{\theta},\sigma_u^2,\sigma_f^2\}} =\frac{\partial \mathcal{L}_\mathcal{NLML}}{\partial  \mathbf{K}_\texttt{DKL}} \frac{\partial \mathbf{K}_\texttt{DKL}}{\partial\{ \boldsymbol{\theta},\sigma_u^2,\sigma_f^2\}} \,.
    \label{eq:deriv_L_theta}
\end{equation}
\paragraph{Enhanced initialization with PDE constraints}\label{initialization}
The marginal likelihood is the probability that we can obtain our observed dataset $\mathcal{D}_{uf}$ by integrating over all possible functions according to the GP prior. Poor initialisation can cause models to be located in unwanted local minima \cite{jia2021physics}. Moreover, \cite{lotfi2022bayesian} claimed the NLML optimisation can overfit and is sensitive to prior assumptions. However, we find that if physical knowledge is leveraged to help inform the initialisation of kernel hyperparameters $\boldsymbol{\theta}=\left\{\sigma^2, \ell\right\}$, the subsequent model training can not only be accelerated but also generalise well with a reasonable amount of training samples.

Specifically, some modifications will be made to the existing data to realise improved initialisation. Firstly, we use the variance of the $u-$data set as the $\sigma_{\text{temp}}^2$ for our pre-initialising kernel, then, we `project' the $f-$data onto $u$, equivalent to performing a GP approximation at the locations ${\vb{Q}}_f$, obtaining a new $\tilde{\mathbf{y}}_{f \rightarrow u}$ given by $\mathbf{y}_f$ \eqref{eq:init_sigma}. Here $\sigma_{\text{temp}}^2$ is used in the kernel in the projection step. Subsequently, we will combine the newly obtained $\tilde{\mathbf{y}}_{f \rightarrow u}$ and $\mathbf{y}_u$ together to form a new data vector $\tilde{\mathbf{y}}_{uf}$. The variance of $\tilde{\mathbf{y}}_{uf} =\binom{ \mathbf{y}_u}{ \tilde{\mathbf{y}}_{f \rightarrow u} } $  will be utilized and serve as the initialized value for $\sigma^2$. For the length scales, we compute the distances across all data samples and then use the average values of these distances as the initial values. Algorithm \ref{alg:init} presents the detailed steps of the proposed method.
\begin{equation}
    \tilde{\mathbf{y}}_{f \rightarrow u} = \left(\mathcal{A}_{\vb*{q}'} k\left({\vb{Q}}_f, {\vb{Q}}_f\right)\right) \left(\mathcal{A}_{\vb*{q}} \mathcal{A}_{\vb*{q}^{\prime}} k\left({\vb{Q}}_f, {\vb{Q}}_f\right)\right) ^{-1}\mathbf{y}_f \,.
    \label{eq:init_sigma}
\end{equation}

\begin{algorithm}
\caption{Bayesian optimisation for final layer neurons in DKL model}
\label{alg:bayesop}
\begin{algorithmic}[1] 
\Ensure 
The optimal number of neurons in the final layer $n_h^{L*}$
\Procedure{BayesOptDKL}
{input data : $\mathbf{Q}_u$, $\mathbf{Q}_f$, $\mathbf{y}_u$, $\mathbf{y}_f$, initial network architecture settings: $\mathcal{NN}_{\text{base}}$, search set: $\mathcal{S}$, total number of evaluations $N_{\text{eval}}$,
acquisition function $\alpha$, number of initial random evaluations $n_{\text{init}}$}      
    \State initialize $\mathcal{L}_{\mathcal{NLML}}^* \gets \infty$, $n_h^{L*} \gets \textit{null}$ 
        \Comment{\textcolor{gray}{Initialize best loss and neuron number}}
    \State randomly select $n_{\text{init}}$ values $\{ n_h^{L(i)} \}_{i=1}^{n_{\text{init}}} \subset \mathcal{S}$
    \State initialize an empty dataset $\mathcal{D} \gets \emptyset$

    \Function{Objective}{$n_h^L$}
        \State Update $\mathcal{NN}$ with $n_h^L$
        \Comment{\textcolor{gray}{Configure $\mathcal{NN}$ to have $n_h^L$ neurons in the last layer}}
    
        \State $\mathbf{Z}_u \leftarrow \mathcal{NN}(\mathbf{Q}_u)$, $\mathbf{Z}_f \leftarrow \mathcal{NN}(\mathbf{Q}_f)$
        \Comment{\textcolor{gray}{Compute embeddings}}
    
        \State
        $
        \mathbf{K}_{\texttt{DKL}} \leftarrow \begin{pmatrix}
            k(\mathbf{Z}_u, \mathbf{Z}_u) + \sigma_u^2 \mathbf{I} & k(\mathbf{Z}_u, \mathbf{Z}_f) \\
            k(\mathbf{Z}_f, \mathbf{Z}_u) & k(\mathbf{Z}_f, \mathbf{Z}_f) + \sigma_f^2 \mathbf{I}
        \end{pmatrix}
        $
        , $\mathbf{y} \leftarrow \begin{pmatrix} \mathbf{y}_u \\ \mathbf{y}_f \end{pmatrix}$
        \State 
                    $\mathcal{L}_{\mathcal{NLML}} \gets \frac{1}{2} \mathbf{y}^\top \mathbf{K}_\texttt{DKL}^{-1} \mathbf{y} + \frac{1}{2} \log |\mathbf{K}_\texttt{DKL}| + \frac{N+M}{2} \log (2\pi)$
        \State \Return $\mathcal{L}_{\mathcal{NLML}}$
    \EndFunction

    \For{$i = 1$ to $n_{\text{init}}$}
        \State $\mathcal{L}_{\mathcal{NLML}}^{(i)} \gets \Call{Objective}{n_h^{L(i)}}$
        \Comment{\textcolor{gray}{Compute negative log marginal likelihood for each initial $n_h^L$}}
        \State Update dataset $\mathcal{D} \gets \mathcal{D} \cup \{ (n_h^{L(i)}, \mathcal{L}_\mathcal{NLML}^{(i)}) \}$
        \If{$\mathcal{L}_{\mathcal{NLML}}^{(i)} < \mathcal{L}_{\mathcal{NLML}}^{*}$}
            \State $\mathcal{L}_{\mathcal{NLML}}^{*} \gets \mathcal{L}_{\mathcal{NLML}}^{(i)}$
            \State $n_h^{L*} \gets n_h^{L(i)}$
        \EndIf
    \EndFor

    \For{$t = n_{\text{init}} + 1$ to $N_{\text{eval}}$}
        \State train a GP surrogate model $\mathcal{GP}$ using data $\mathcal{D}$
        \State select next $n_h^L$ by maximising the acquisition function:
        $
        n_h^{L(t)} = \arg\max_{n_h^L \in \mathcal{S}} \alpha(n_h^L \mid \mathcal{D})
        $
        \State $\mathcal{L}_{\mathcal{NLML}}^{(t)} \gets \Call{Objective}{n_h^{L(t)}}$
        \Comment{\textcolor{gray}{Evaluate objective at selected $n_h^L$}}
        \State Update dataset $\mathcal{D} \gets \mathcal{D} \cup \{ (n_h^{L(t)}, \mathcal{L}_{\mathcal{NLML}}^{(t)}) \}$
        \If{$\mathcal{L}_{\mathcal{NLML}}^{(t)} < \mathcal{L}_{\mathcal{NLML}}^{*}$}
            \State $\mathcal{L}_{\mathcal{NLML}}^{*} \gets \mathcal{L}_{\mathcal{NLML}}^{(t)}$
            \State $n_h^{L*} \gets n_h^{L(t)}$
        \EndIf
    \EndFor
    \State \Return $n_h^{L*}$
\EndProcedure 
\end{algorithmic} 
\end{algorithm}

\begin{algorithm}
\caption{Parameter Initialization for $\{\boldsymbol{\omega}, \sigma^2, \boldsymbol{\ell}\}$}\label{alg:init}
\begin{algorithmic}[1] 
\Ensure
Initialized hyperparameters: $\boldsymbol{\omega}, \sigma^2, \boldsymbol{\ell}$

\Procedure{ParamsInit}{$\mathbf{Q}_u, \mathbf{Q}_f, \mathbf{y}_u, \mathbf{y}_f, \mathcal{A}, \mathcal{NN}$}
    \State $\sigma_{\text{temp}}^2 \gets \text{var}(\mathbf{y}_u)$ \Comment{\textcolor{gray}{Estimate the initial variance from $\mathbf{y}_u$}}
    \State $\tilde{\mathbf{y}}_{f \rightarrow u} \gets \left(k\left(\mathbf{Q}_f, \mathbf{Q}_f\right)\mathcal{A}^\top \right)^{-1} \left(\mathcal{A} k\left(\mathbf{Q}_f, \mathbf{Q}_f\right)\mathcal{A}^\top \right) \mathbf{y}_f$ \Comment{\textcolor{gray}{Project the observed points of $f$ onto $u$ through a GP approximation}}
    \State $\tilde{\mathbf{y}}_{uf} \gets [\mathbf{y}_u; \tilde{\mathbf{y}}_{f \rightarrow u}]$ \Comment{\textcolor{gray}{Concatenate the observed and predicted outputs}}
    \State $\sigma^2 \gets \text{var}(\tilde{\mathbf{y}}_{uf})$ \Comment{\textcolor{gray}{Recalculate the variance from the combined outputs}}
    \State $\mathbf{Q} \gets [\mathbf{Q}_u; \mathbf{Q}_f]$ \Comment{\textcolor{gray}{Concatenate input points from $\mathbf{Q}_u$ and $\mathbf{Q}_f$}}
    \State $\boldsymbol{\ell} \gets \frac{1}{\binom{N}{2}} \sum_{i=1}^{N-1} \sum_{j=i+1}^{N} \sqrt{ (\vb*{q}^{(i)} - \vb*{q}^{(j)})^2}, \quad \text{for all } i, j \in \{1, \dots, N\}$
    \Comment{\textcolor{gray}{Compute the average Euclidean distance between all unique pairs of points $(\vb*{q}^{(i)},\vb*{q}^{(j)})$ collected in $\mathbf{Q}$ to initialise $\boldsymbol{\ell}$, {$N$} being the total number of points in $\mathbf{Q}$}}

    \State $\boldsymbol{\omega} \gets \Call{LecunNorm}{\mathcal{NN}}$ \Comment{\textcolor{gray}{Initialize using the LeCun normal to maintain the variance of activations across layers \cite{klambauer2017self}}}
    \State \textbf{return} $\boldsymbol{\omega}, \sigma^2, \boldsymbol{\ell}$
\EndProcedure
\end{algorithmic}
\end{algorithm}
\begin{algorithm}
\caption{optimisation of trainable parameters with PDE constraints}\label{alg:optimisation}
\begin{algorithmic}[1]
\Ensure
Optimized parameters: $\boldsymbol{\omega}$, $\boldsymbol{\theta}$, $\sigma_u^2$, $\sigma_f^2$

\Procedure{OptimizeWithPDEs}{input data: $\mathbf{Q}_u, \mathbf{Q}_f, \mathbf{y}_u, \mathbf{y}_f$, differential operator: $\mathcal{A}$, network configuration: $\mathcal{NN}$, maximum iterations: $T$, convergence threshold: $\epsilon$}
    \State $\boldsymbol{\omega}, \boldsymbol{\theta} \gets \Call{ParamsInit}{\mathbf{Q}_u, \mathbf{Q}_f, \mathbf{y}_u, \mathbf{y}_f, \mathcal{A}, \mathcal{NN}}$ 
    \Comment{\textcolor{gray}{see algorithm(\ref{alg:init})}}
    \State $\mathbf{y} \gets \begin{pmatrix} \mathbf{y}_u \\ \mathbf{y}_f \end{pmatrix}$
    \State $t \gets 0$, $\delta \gets \infty$ 
    \Comment{\textcolor{gray}{set iteration count and change}}
    \Repeat
        \State $\mathbf{K}^{1} \gets k_\texttt{DKL}({\mathbf{Q}}_u,{\mathbf{Q}}_u)+\sigma_u^2 \mathbf{I}$
        \State $\mathbf{K}^{2} \gets k_\texttt{DKL}\left({\mathbf{Q}}_u,{\mathbf{Q}}_f\right)\mathcal{A}^\top$
        \State $\mathbf{K}^{3} \gets \mathcal{A} k_\texttt{DKL}\left({\mathbf{Q}}_f,{\mathbf{Q}}_u\right)$
        \State $\mathbf{K}^{4} \gets \mathcal{A} k_\texttt{DKL}\left({\mathbf{Q}}_f,{\mathbf{Q}}_f\right)\mathcal{A}^\top+{\sigma_f^2 \mathbf{I}}$
        \Comment{\textcolor{gray}{compute all kernel function mappings for a given operator}}
        \State  $
                \mathbf{K}_\texttt{DKL} \gets
                \begin{pmatrix}
                \mathbf{K}^{1} & \mathbf{K}^{2} \\
                \mathbf{K}^{3} & \mathbf{K}^{4}
                \end{pmatrix}
                $

        \State $\mathcal{L}_{\mathcal{NLML}} \gets \frac{1}{2} \mathbf{y}^\top \mathbf{K}_\texttt{DKL}^{-1} \mathbf{y} + \frac{1}{2} \log |\mathbf{K}_\texttt{DKL}| + \frac{N+M}{2} \log (2\pi)$
        \Comment{\textcolor{gray}{optimisation objective}}
        \State 
        $\boldsymbol{\omega} \gets \boldsymbol{\omega} - \eta_{\boldsymbol{\omega}} \nabla_{\boldsymbol{\omega}} \mathcal{L}_{\mathrm{NLML}}^{(t)}, \quad
        \boldsymbol{\theta} \gets \boldsymbol{\theta} - \eta_{\boldsymbol{\theta}} \nabla_{\boldsymbol{\theta}} \mathcal{L}_{\mathrm{NLML}}^{(t)}, \quad
        \sigma_u^2 \gets \sigma_u^2 - \eta_{\sigma_u^2} \frac{\partial \mathcal{L}_{\mathrm{NLML}}^{(t)}}{\partial \sigma_u^2}, \quad
        \sigma_f^2 \gets \sigma_f^2 - \eta_{\sigma_f^2} \frac{\partial \mathcal{L}_{\mathrm{NLML}}^{(t)}}{\partial \sigma_f^2}$
        \State $\delta \gets |\mathcal{L}_{\mathcal{NLML}}^{(t)} - \mathcal{L}_{\mathcal{NLML}}^{(t-1)}|$
        \State $t \gets t + 1$
    \Until{$\delta < \epsilon$ \textbf{or} $t \geq T$}
    \State \textbf{return} $\boldsymbol{\omega}, \boldsymbol{\theta}, \sigma_u^2, \sigma_f^2$
\EndProcedure
\end{algorithmic}
\end{algorithm}

\section{Numerical Results}
In this section, we present numerical results across four benchmark problems to demonstrate the effectiveness of the proposed method in solving high-dimensional forward PDE problems while simultaneously providing meaningful uncertainty quantification. In particular, we compare our approach against the PDE-constrained GP method \cite{raissi2017machine} (i.e., without integrating NNs). Our results show that the proposed method can address PDE problems in dimensions as high as 50, far beyond the capability of classical numerical discretisation and the PDE-constrained GPs with simple kernels. This section provides a comprehensive comparison, emphasising our approach's accuracy, computational efficiency, and uncertainty quantification capabilities, particularly in high-dimensional settings where traditional methods face significant limitations.

\paragraph{Experimental Setup}
In this study, we solve four high-dimensional PDEs: the parameterised heat equation, Poisson's, advection-diffusion-reaction, and heat equations.
Due to the challenge of acquiring real high-dimensional data, we generate simulated data to evaluate the proposed method. The dataset includes boundary points, initial points, and randomly sampled points within the domain. Analytical solutions are used as the ground truth for evaluating model accuracy.

We differentiate between hyperparameters for NN-related $\boldsymbol{\omega}$ and GP-related  $\boldsymbol{\theta}$. The number of NN-related parameters varies with the network size, where weights are initialised using the LeCun normal method and biases are initialised to zero. GP-related hyperparameters consist of variance, lengthscales, and noise variances, where the number of lengthscales for the PDE-GP model equals the input dimension $d$. For the PDE-DKL model, it equals the number of neurons in the final NN layer. 

The NN architecture consists of four layers with a width of 200 neurons each, which balances expressiveness and optimisation complexity. GELU is selected as the activation function due to its smooth derivatives, ensuring stable gradients and faster convergence, which is crucial for kernel computation. 
A warm-up phase and learning rate annealing strategy are incorporated to accelerate training and enhance generalisation. 
We employ random sampling techniques and gradient clipping to maintain stability during the optimisation process, which uses the Adam optimiser with a moderate learning rate and exponential decay. 
The \verb+jax+ and \verb+optax+ libraries are used for data generation and optimisation. 

All computational experiments were completed on a \verb+Dell R750xa+ with \verb+NVIDIA L40/48G+ GPUs. Our analysis focuses on test accuracy, measured by the relative $L_2$ error against the ground truth, and computational efficiency, considering both time and memory consumption. Furthermore, the model's predictive accuracy and uncertainty quantification will be compared against the analytical solutions to assess the method's capabilities comprehensively.

\subsection{Parametric heat equation}
In this section, we consider the following parametric, time-dependent heat equation with boundary and initial conditions:
\begin{equation}
\begin{cases}
\partial_t u(x, t, \vb*{\mu})-\frac{\partial^2 u(x, t, \vb*{\mu})}{\partial x^2} =f\left(x, t, \vb*{\mu}\right) & \text { in } \Omega \times[0, T] \,,\\ 
u(0, t,\vb*{\mu})=0  & \text { in } [0, T] \,,\\
u(1, t,\vb*{\mu})=\mu_3 \mathrm{e}^{-\mu_1 t} \sin \left(2 \pi \mu_2\right) & \text { in } [0, T] \,,\\
u(x, 0,\vb*{\mu})=\mu_3 \sin \left(2 \pi \mu_2 x\right)  & \text { in } \Omega \,,
\end{cases}
\label{eq:para_heat}
\end{equation}
where $\Omega=(0,1) \subset \mathbb{R}$ is the spatial domain, $T=1$ is the final time, and the parameter domain is given by $\vb*{\mu} = \{\mu_1,\mu_2,\mu_3\}^\top\in \mathcal{M} = [0.8,1.2] \times [0.7,1.3] \times [0.9,1.1]$. 
The forcing term is defined to be
$
f\left(x, t, \vb*{\mu}\right)=\mu_3 \mathrm{e}^{-\mu_1 t} \sin \left(2 \pi \mu_2 x\right)\left(-\mu_1+4 \pi^2 \mu_2^2\right)
$. In this case, the exact solution to this initial boundary value problem is 
$
u\left(x, t, \vb*{\mu}\right)=\mu_3 \mathrm{e}^{-\mu_1 t} \sin \left(2 \pi \mu_2 x\right)
$.

\begin{figure}[htbp]
    \centering
    \includegraphics[width=\textwidth]{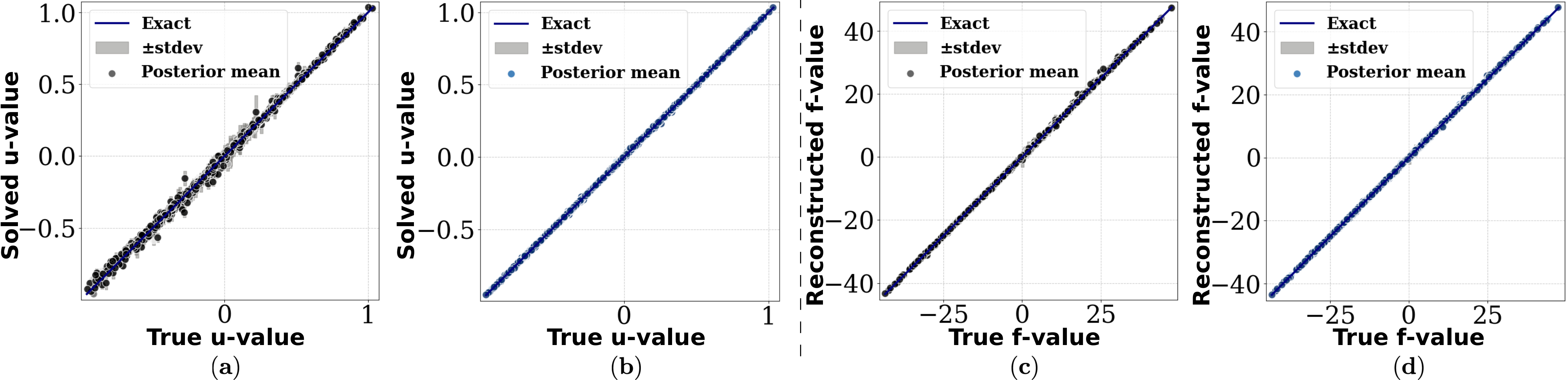}
    \caption{Comparison of PDE-GP (black dots) and PDE-DKL (blue dots) in predicting the solution term $u$ and reconstructing the forcing term $f$. Subplots ($\textbf{a}$) and (\textbf{b}) show the predictions for $u$ using GP and DKL, respectively, while subplots (\textbf{c}) and (\textbf{d}) show the reconstructions for $f$ using GP and DKL respectively.}
    \label{fig:para}
\end{figure}

To evaluate the proposed model's performance, we apply the PDE-constrained Gaussian process (PDE-GP) and PDE-constrained deep kernel learning (PDE-DKL) models, comparing their capabilities in predicting the solution term $u$ and reconstructing the forcing term $f$. Figure~\ref{fig:para} illustrates the performance of both models. 
The results demonstrate that both models can effectively predict the solution of the parametric heat equation with reasonable accuracy. However, the PDE-DKL shows an improved prediction accuracy due to its enhanced expressive capacity, allowing it to better model the complexity introduced by the parameters $\vb*{\mu}$. The PDE-DKL model outperforms the PDE-GP under conditions of more significant parametric variability, as shown in subplots (\textbf{b}) and (\textbf{d}), where the DKL provides a more accurate recovery of $u$ and reconstruction of $f$.
This highlights the potential of the PDE-DKL approach in effectively handling complex, parameter-dependent PDEs.

\subsection{High-dimensional Poisson equation}
In this section, we evaluate the performance of the proposed method by solving the high-dimensional Poisson equation, which is fundamental in classical mathematical physics and plays a critical role in modelling various phenomena across physics, engineering, and data science. We focus on dimensions $d=10$ and $d=50$ to assess the method's potential in high-dimensional settings. The Poisson equation is formulated as
\begin{equation}
    \begin{cases}
    \Delta u(\boldsymbol{x})=f(\boldsymbol{x})  & \text { in } \Omega=(0,1)^d \,,\\ u(\boldsymbol{x})=g(\boldsymbol{x}) & \text { on } \partial \Omega \,,
\end{cases} \label{eq:high_poisson}
\end{equation}
where $\boldsymbol{x} = (x_1, x_2, \dots, x_d)^\top \in \mathbb{R}^d$ spans a $d$ -dimensional hypercube domain $\Omega = [0,1]^d$, and $\Delta$ denotes the Laplacian operator in $d$ dimensions.
In this test, we set the source term $f(\boldsymbol{x})$ and the boundary condition $g(\boldsymbol{x})$ as follows:
$
    f(x)=-d \sin \left(\mathbf{1}^{\top} \boldsymbol{x}\right)-d \cos \left(\mathbf{1}^{\top} x\right) \text { for } x \in \Omega=[0,1]^d
$, and
$
    g(\boldsymbol{x})=\sin \left(\mathbf{1}^{\top} \boldsymbol{x}\right)+\cos \left(\mathbf{1}^{\top} \boldsymbol{x}\right)\text { for } \boldsymbol{x} \in \partial \Omega
$. Here, $\mathbf{1}$ denotes a vector whose entries are all 1.

This choice of $f(\boldsymbol{x})$ ensures that the solution exhibits non-trivial behaviour across all dimensions.
In this case, the exact solution of the Poisson equation \eqref{eq:high_poisson} is:
$
    u(\boldsymbol{x}) = \sin\left(\mathbf{1}^\top \boldsymbol{x}\right) + \cos\left(\mathbf{1}^\top \boldsymbol{x}\right).
    \label{eq:exact_solution}
$

\begin{figure}[htbp]
    \centering
        \includegraphics[width=\textwidth]{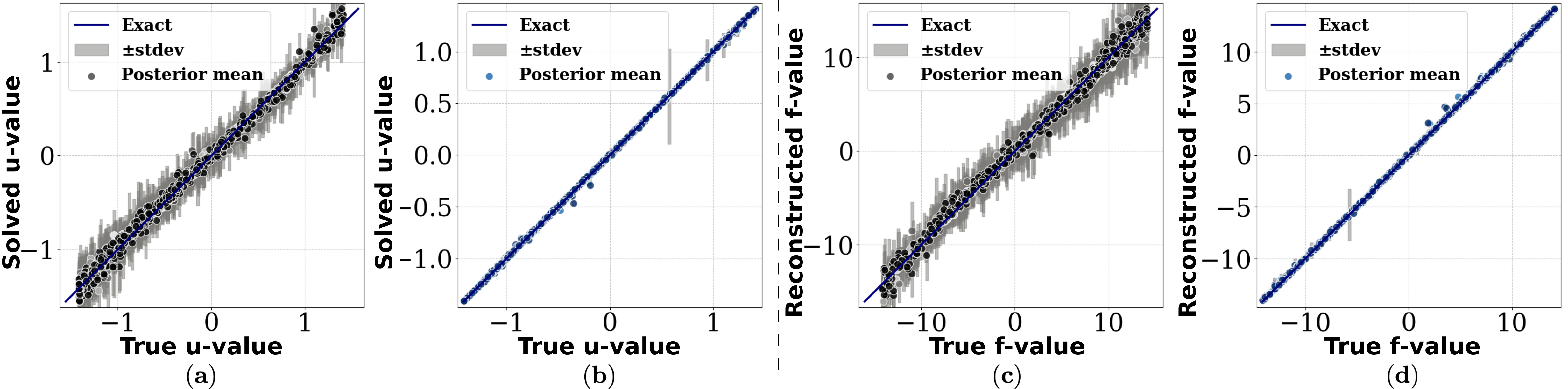}
        \label{fig:compare-poisson_10}
    \caption{Comparison of PDE-GP and PDE-DKL methods in solving the ten-dimensional Poisson equation. Subplots  (\textbf{a}) and  (\textbf{b}) show the predicted solutions for $u$ using PDE-GP and PDE-DKL, respectively. Subplots  (\textbf{c}) and  (\textbf{d}) display the reconstructed source term $f$ using PDE-GP and PDE-DKL, respectively.
    }
    \label{fig:poisson_10}
\end{figure}
Figure \ref{fig:poisson_10} compares the performance of PDE-GP and PDE-DKL in solving the ten-dimensional Poisson equation. Subplots ($\textbf{a}$) and  (\textbf{b}) depict scatter plots of the predicted values of $u$ versus the true values for PDE-GP and PDE-DKL, respectively. Subplots (\textbf{c}) and  (\textbf{d}) illustrate the reconstruction of the source term $f$ by both methods. Ideal prediction accuracy corresponds to points lying along the diagonal line. Error bars represent the standard deviation, indicating the prediction uncertainty.

\begin{figure}[!htb]
    \centering
    \includegraphics[width=0.6\textwidth]{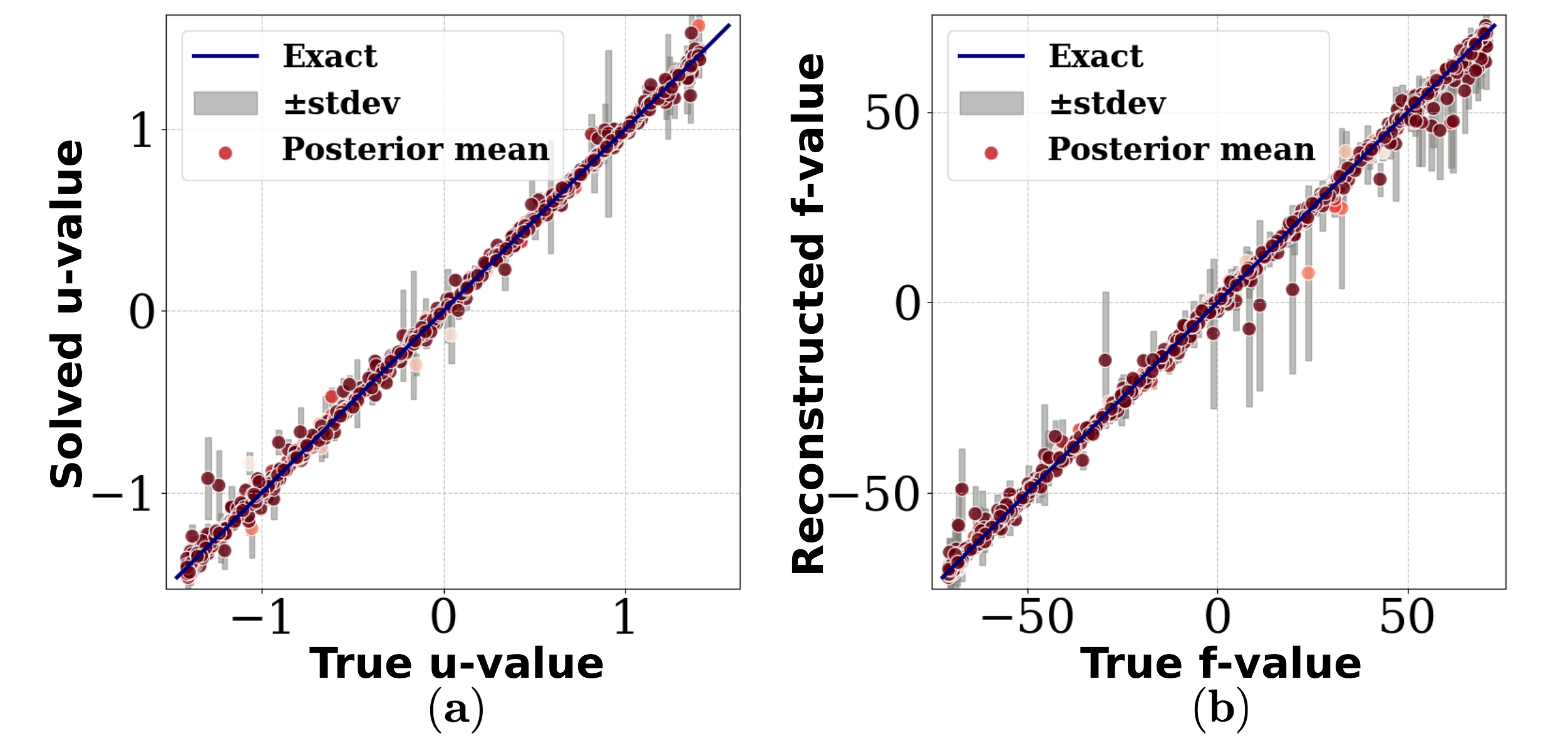}
    \caption{Accuracy of PDE-DKL and PDE-GP in solving high-dimensional Poisson equations. Subfigure ($\textbf{a}$) details the accuracy and uncertainty of PDE-DKL in solving for variable $u$ in a 50-dimensional Poisson equation. Subfigure ($\textbf{b}$) illustrates the accuracy and uncertainty in reconstructing $f$ using PDE-DKL, demonstrating its superior performance and efficiency in handling large-scale computations.}
\label{fig:poisson_50}
\end{figure}
The results demonstrate that PDE-DKL outperforms PDE-GP in predicting $u$ and reconstructing $f$ for the ten-dimensional Poisson equation (Figure \ref{fig:poisson_10}). PDE-DKL's predictions align closely with the ideal diagonal line, indicating higher accuracy and reduced variance than PDE-GP. Despite using identical training dataset sizes, PDE-DKL achieves superior prediction accuracy while significantly reducing computational requirements. This efficiency makes PDE-DKL preferable for high-dimensional applications with limited data availability. Additionally, from a practical point of view, PDE-GP requires substantially more memory due to the need to compute and store large covariance matrices. In contrast, PDE-DKL efficiently handles high-dimensional data with reduced memory consumption.

Figure \ref{fig:poisson_50} presents the performance of PDE-DKL in solving the 50-dimensional Poisson equation. Subplot ($\textbf{a}$) shows the predicted values of $u$ versus the true values, while subplot ($\textbf{b}$) displays the reconstructed source term $f$. The scatter plots consist of 1,000 randomly sampled points, with error bars representing the standard deviation. The close alignment of points along the diagonal line indicates that PDE-DKL maintains high prediction accuracy even in high-dimensional settings. PDE-GP was not included in this comparison due to computational limitations in handling the 50-dimensional case.
As the dimensionality increases to $d=50$, PDE-GP becomes computationally infeasible due to the substantial memory required to compute and store large covariance matrices and their derivatives. Specifically, memory consumption grows rapidly with dimension, making it impractical for standard computational resources. In contrast, PDE-DKL effectively handles high-dimensional data by leveraging deep kernel learning, which reduces the need to store and invert large matrices. Despite the high dimensionality and limited training data, PDE-DKL achieves accurate predictions with manageable computational resources. This demonstrates the scalability and efficiency of PDE-DKL in solving high-dimensional PDEs.

\subsection{High-dimensional heat equation}
We extend our analysis to time-dependent problems to further evaluate the proposed model by considering the high-dimensional heat equation in spatial dimensions $d=10$ and $d=50$. The equation under consideration is given by:
\begin{equation}
    \begin{cases}
        \partial_t u(\boldsymbol{x}, t) - \Delta u(\boldsymbol{x}, t) = f(\boldsymbol{x}, t) & \text{in } \Omega \times [0, T]\,, \\
        u(\boldsymbol{x}, t) = g(\boldsymbol{x}, t) & \text{on } \partial \Omega \times [0, T]\,, \\
        u(\boldsymbol{x}, 0) = h(\boldsymbol{x}) & \text{in } \Omega \,,
    \end{cases}
    \label{eq:heat_equation}
\end{equation}
where 
$T=1$ represents the final time.
The source term is defined as $\mathcal{A}_{\vb*{q}} [u({\vb*{q}})] \equiv f(x, t)=$ $\left(-1+\frac{1}{d}\right) \mathrm{e}^{-t} \cos \left(\frac{1}{d} \mathbf{1}^{\top} \boldsymbol{x}\right)$, the boundary condition as $g(\boldsymbol{x}, t)=\mathrm{e}^{-t} \cos \left(\frac{1}{d} \mathbf{1}^{\top} \boldsymbol{x}\right)$, and the initial condition as $h(\boldsymbol{x})=\cos \left(\frac{1}{d} \mathbf{1}^{\top} \boldsymbol{x}\right)$.
To satisfy the heat equation \eqref{eq:heat_equation}, along with the specified source term, boundary, and initial conditions, the exact solution is given by $u(\boldsymbol{x}, t)=$ $\mathrm{e}^{-t} \cos \left(\frac{1}{d} \mathbf{1}^{\top} \boldsymbol{x}\right)$.

\begin{figure}[htbp]
    \centering
        \includegraphics[width=\textwidth]{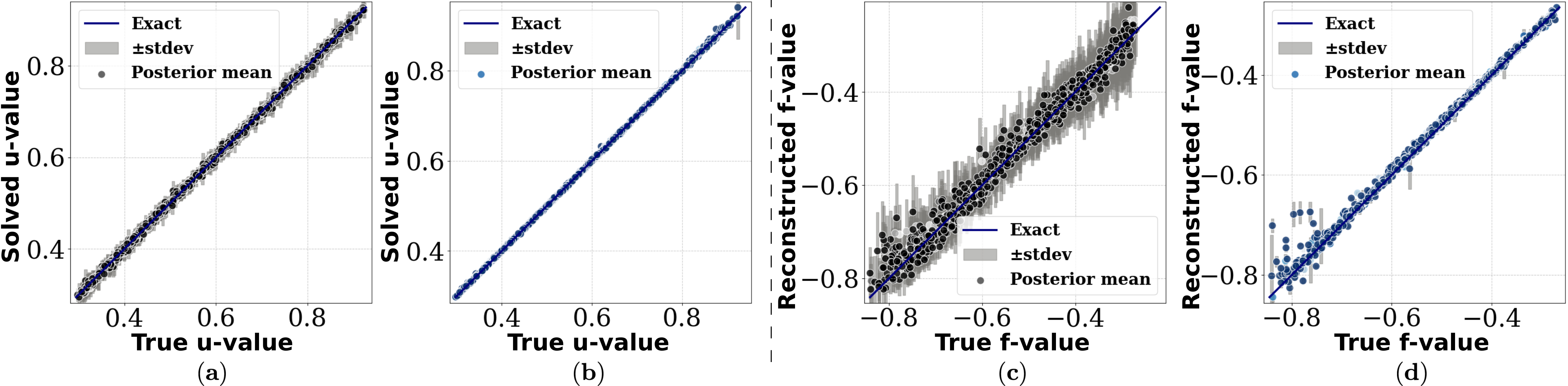}
        \caption{Subfigures (\textbf{a}) and (\textbf{b}) show scatter plots of predicted versus true values of $u$ for PDE-GP and PDE-DKL, respectively. 
        Subplots (\textbf{c}) and (\textbf{d}) display the reconstructed source term $f$ using PDE-GP and PDE-DKL, respectively.
        }
        \label{fig:heat_10}
\end{figure}

The results in Figure \ref{fig:heat_10} indicate that PDE-DKL provides more accurate predictions for $u$ and $f$ than PDE-GP in the ten-dimensional case. Specifically, the predictions from PDE-DKL align more closely with the true values, as evidenced by the clustering of points along the diagonal line in subplots (b) and (d). Additionally, the error bars for PDE-DKL are smaller, indicating lower uncertainty in the predictions. In contrast, PDE-GP exhibits larger variances and less accurate predictions, suggesting limitations in scalability to higher dimensions.

\begin{figure}[!htb]
    \centering
        \includegraphics[width=0.6\textwidth]{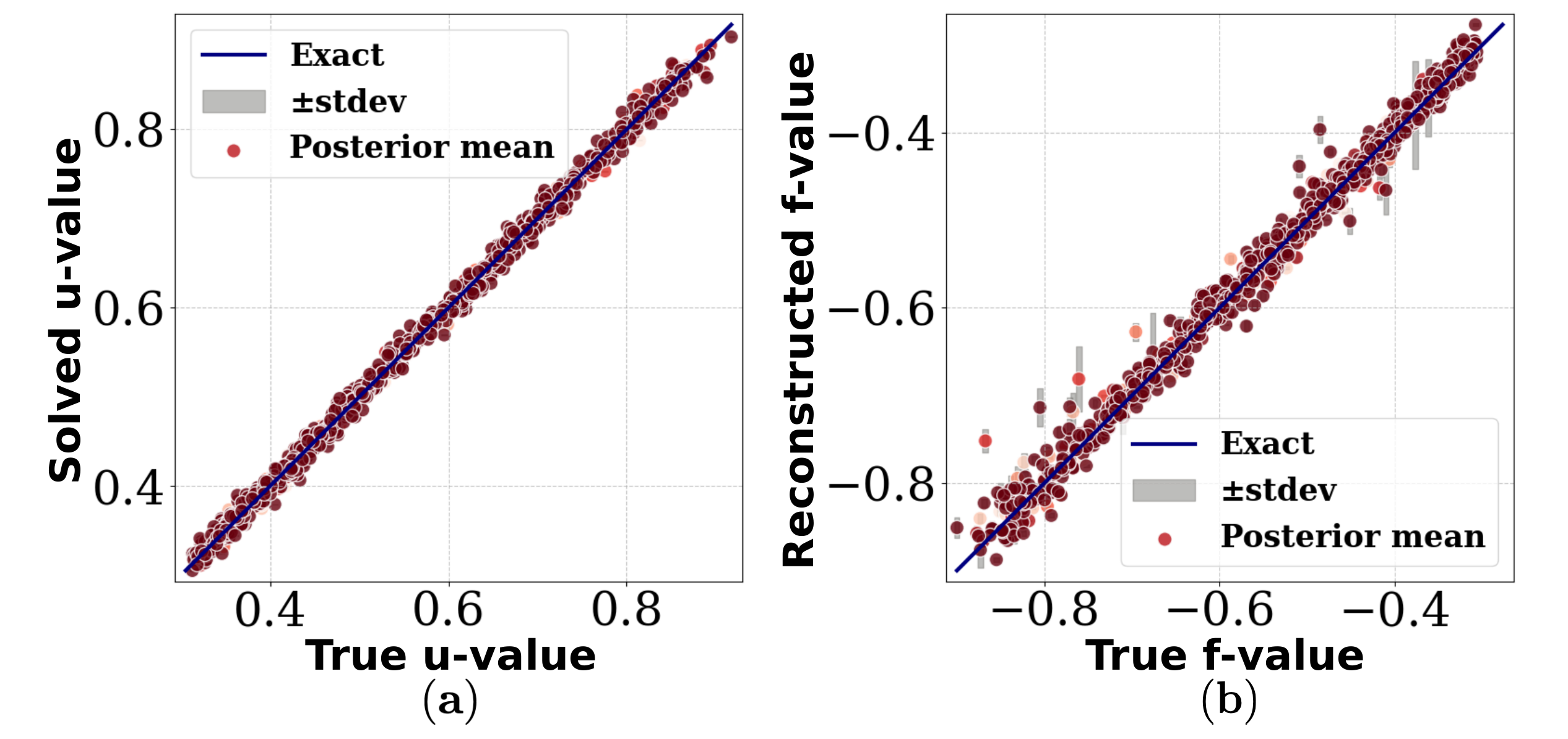}
    \caption{Predictive accuracy of PDE-DKL versus PDE-GP in solving the high-dimensional heat equation. Subfigure (\textbf{a}) examines the precision and reliability of PDE-DKL in determining the solution for variable $u$ in a 50-dimensional setting. Subfigure (\textbf{b}) illustrates the accuracy and uncertainty in reconstructing the source term $f$ using PDE-DKL, demonstrating its enhanced capability to manage complex calculations with minimal uncertainty.}
        \label{fig:heat_50}
\end{figure}
Figure \ref{fig:heat_50} illustrates the performance of PDE-DKL in solving the 50-dimensional heat equation. Subplot (\textbf{a}) shows the predicted versus true values of $u$, and subplot (\textbf{b}) presents the reconstruction of the source term $f$. Each scatter plot contains 1,000 randomly sampled points, with error bars representing prediction uncertainty. Due to computational limitations, PDE-GP results are not presented for this high-dimensional case.
The results in Figure \ref{fig:heat_50} demonstrate that PDE-DKL effectively solves the 50-dimensional heat equation with high accuracy and low uncertainty. The predicted values of $u$ and reconstructed values of $f$ closely match the true values, as shown by the alignment of points along the diagonal line. The small error bars indicate that the method provides reliable uncertainty quantification. PDE-GP was not applied in this case because of its excessive computational burden in high dimensionality, highlighting PDE-DKL's superior scalability and efficiency for solving high-dimensional PDEs.

\subsection{High-dimensional advection-diffusion-reaction equation}
In addition, we evaluate the proposed model using the high-dimensional advection-diffusion-reaction equation. This equation incorporates additional complexities by accounting for both advection and reaction processes.
The equation is formulated as:
\begin{equation}
    \begin{cases}
        \partial_t u(\boldsymbol{x}, t)-\Delta u(\boldsymbol{x}, t)+\mathbf{1}^{\top}  \nabla u(\boldsymbol{x}, t)+u(\boldsymbol{x}, t)=f(\boldsymbol{x}, t) & \text { in } \Omega \times[0, 1] \,, \\ 
        u(\boldsymbol{x}, t)=g(\boldsymbol{x}, t) & \text { on } \partial \Omega \times[0, 1] \,, \\ 
        u(\boldsymbol{x}, 0)=h(\boldsymbol{x}) & \text { in } \Omega \,.
    \end{cases}
    \label{eq:advect_r_a}
\end{equation}
The spatial domain is defined as $\Omega=(0,1)^d \subset \mathbb{R}^d$, with two test cases considered for $d=10$ and $d=50$. The source term is given by: $\mathcal{A}_{\vb*{q}} [u({\vb*{q}})] \equiv f(\vb*{x}, t) =e^{-t}\left(\cos \left(\frac{1}{d} \mathbf{1}^{\top}\boldsymbol{x}\right) + \frac{1}{d}\sin \left(\frac{1}{d} \mathbf{1}^{\top}\boldsymbol{x}\right)\right)$, the boundary condition is $g(\boldsymbol{x}, t)=\mathrm{e}^{-t} \sin \left(\frac{1}{d} \mathbf{1}^{\top} \boldsymbol{x}\right)$, and the initial condition is $h(\boldsymbol{x})=\sin \left(\frac{1}{d} \mathbf{1}^{\top} \boldsymbol{x}\right)$.
The exact solution that satisfies the equation \eqref{eq:advect_r_a}  under these conditions is: $u(\vb*{x}, t) =e^{-t}\sin \left(\frac{1}{d} \mathbf{1}^{\top} \boldsymbol{x}\right)$.

\begin{figure}[htbp]
    \centering
\includegraphics[width=\textwidth]{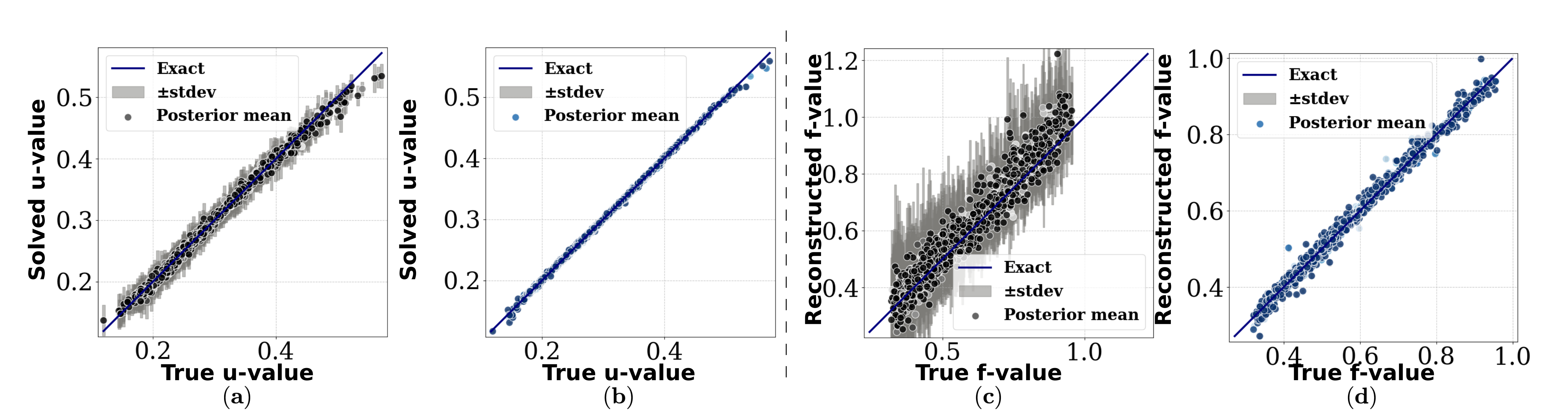}
    \caption{Comparison of PDE-GP and PDE-DKL in solving the ten-dimensional advection-diffusion-reaction equation. Subplots (\textbf{a}) and (\textbf{b}) show predicted versus true values of $u$ for PDE-GP and PDE-DKL, respectively. Subplots (\textbf{c}) and (\textbf{d}) display the source term $f$ reconstruction using PDE-GP and PDE-DKL, respectively.}
    \label{fig:advect_10}
\end{figure}

\begin{figure}[!htb]
    \centering
\includegraphics[width=0.6\textwidth]{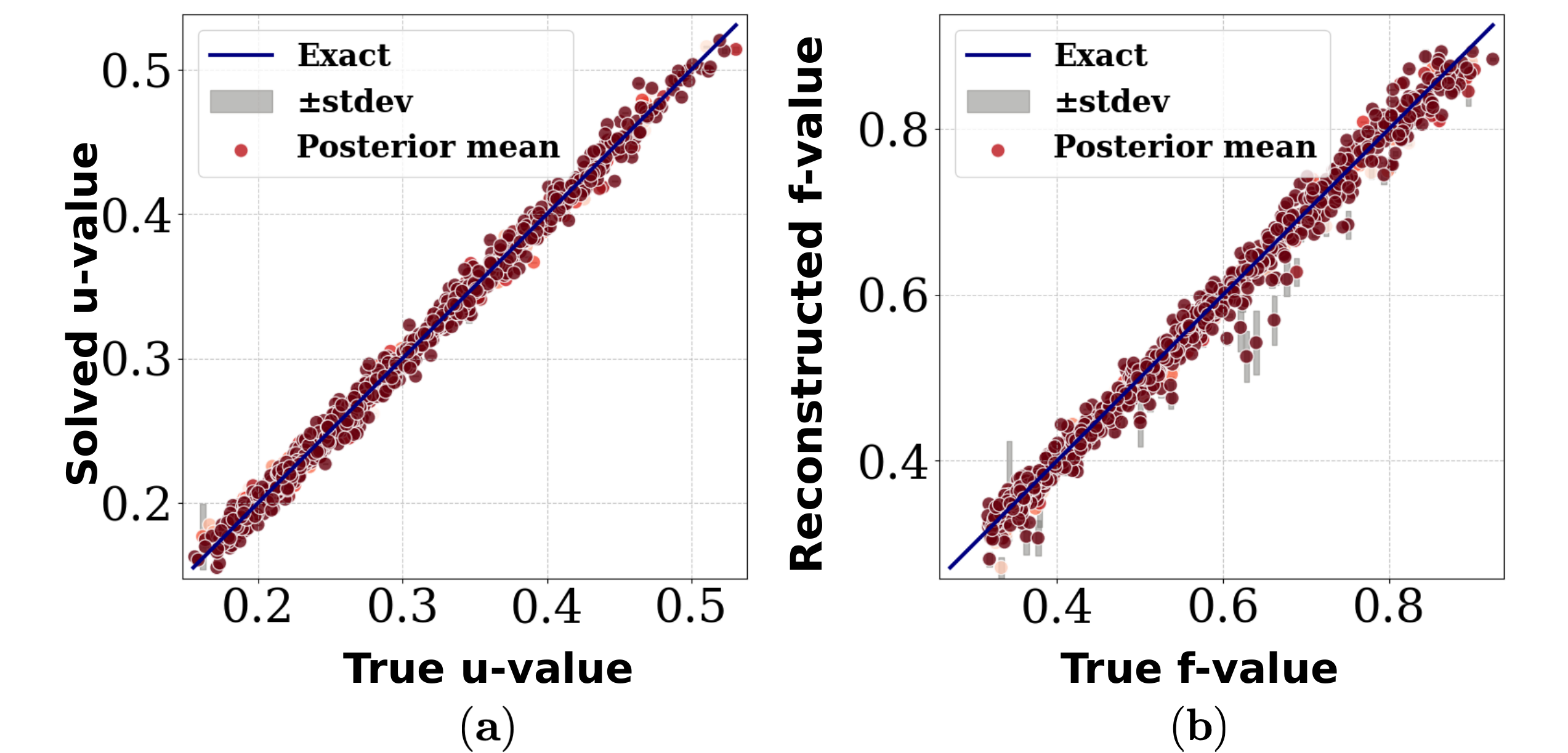}
\caption{Solution fidelity of PDE-DKL versus PDE-GP for the high-dimensional advection-diffusion-reaction equation. Subfigure (\textbf{a}) quantifies the accuracy and consistency of PDE-DKL in resolving the variable $u$ in a 50-dimensional setting. Subfigure (\textbf{b}) illustrates the accuracy and uncertainty in reconstructing the source term $f$ using PDE-DKL, demonstrating its superior performance in conducting intricate simulations with limited uncertainty.}
        \label{fig:advect_50}
\end{figure}

We focus on the PDE-GP and PDE-DKL methods to a 10-dimensional (\ref{fig:advect_10}) and 50-dimensional (\ref{fig:advect_50}) advection-diffusion-reaction equation. 
Subplots (\textbf{a}) and (\textbf{b}) present scatter plots of predicted versus true values of $u$ for PDE-GP and PDE-DKL, respectively. Subplots (\textbf{c}) and (\textbf{d}) show each method's reconstructed source term $f$. Error bars represent the standard deviation, indicating prediction uncertainty.
Figure \ref{fig:advect_50} illustrates the performance of PDE-DKL in solving the fifty-dimensional advection-diffusion-reaction equation. Subplot (\textbf{a}) shows the predicted versus true values of $u$, and subplot (\textbf{b}) presents the reconstruction of the source term $f$. Due to computational constraints, PDE-GP results are not included for this high-dimensional case.
These illustrations detail the performance of PDE-DKL in handling the task, demonstrating its capabilities and effectiveness where PDE-GP falls short, further emphasising PDE-DKL's robustness and adaptability in complex computational scenarios. This figure \eqref{fig:advect_50} evaluates the solution fidelity of PDE-DKL compared to PDE-GP for the high-dimensional advection-diffusion-reaction equation. In subfigure (\textbf{a}), the accuracy and reliability of PDE-DKL in solving for the variable $u$ in a 50-dimensional scenario are examined. Subfigure (\textbf{b}) focuses on the reconstruction accuracy and uncertainty of the source term $f$ using PDE-DKL. The subplots feature scatter plots of 1,000 randomly sampled points, with true values plotted on the $x$-axis and predicted values on the $y$-axis, allowing for a detailed prediction accuracy and consistency assessment. The closer the points are to the diagonal line, the higher the prediction accuracy. Error bars in the plots indicate the standard deviation around each point, providing insights into the prediction uncertainty. 
The results in Figure \ref{fig:advect_50} demonstrate that PDE-DKL handles the fifty-dimensional advection-diffusion-reaction equation with high accuracy and low uncertainty. The predicted values of $u$ and $f$ closely match the true values, as indicated by the alignment along the diagonal line and small error bars. These findings underscore the scalability and robustness of PDE-DKL in solving complex, high-dimensional PDEs where traditional methods like PDE-GP are computationally infeasible.

\begin{table}[htbp]
\centering
\begin{tabular}{l l c c}
\hline
\textbf{PDE (Dimension)} & \textbf{Method} & $\boldsymbol{e}_u$ (\%) & $\boldsymbol{e}_f$ (\%) \\

\hline
Parametric heat equation & PDE-GP  & 4.15\% & 1.26\% \\
                          & PDE-DKL & 0.72\% & 0.82\% \\
\hline
Poisson equation ($d=10$) & PDE-GP  & 6.21\% & 6.38\% \\
                          & PDE-DKL & 0.90\% & 1.06\% \\
Poisson equation ($d=50$) & PDE-GP  & N/A     & N/A     \\
                          & PDE-DKL & 3.38\% & 4.27\% \\
\hline
Heat equation ($d=10$)    & PDE-GP  & 0.78\% & 3.94\%\\
                          & PDE-DKL & 0.29\% & 2.07\% \\
Heat equation ($d=50$)    & PDE-GP  & N/A     & N/A     \\
                          & PDE-DKL & 1.49\% & 2.72\% \\
\hline
Advection-diffusion-reaction ($d=10$) & PDE-GP  & 
2.12\% & 9.10\% \\
                                      & PDE-DKL & 0.71\% & 2.63\% \\
Advection-diffusion-reaction ($d=50$) & PDE-GP  & N/A     & N/A     \\
                                      & PDE-DKL & 2.30\% & 3.01\% \\
\hline
\end{tabular}
\caption{Comparison of relative $L_2$ errors $\boldsymbol{e}_u$ and $\boldsymbol{e}_f$ respectively in solving $ u $ and reconstructing $f$}
\label{tab:l2_errors}
\end{table}
Based on the results presented in Table \ref{tab:l2_errors}, the PDE-DKL method consistently yields lower relative $L_2$ errors than the PDE-GP method in both solving $u$ and reconstructing $f$ across all examined PDEs and dimensions. Specifically, PDE-DKL achieves better accuracy in the 10-dimensional cases and remains computationally feasible and accurate in the 50-dimensional scenarios where PDE-GP is inapplicable. This consistent performance underscores the superior scalability and effectiveness of PDE-DKL for solving high-dimensional PDEs compared to the PDE-GP approach.

\begin{figure}
    \includegraphics[width=\linewidth,height=\textheight,keepaspectratio]{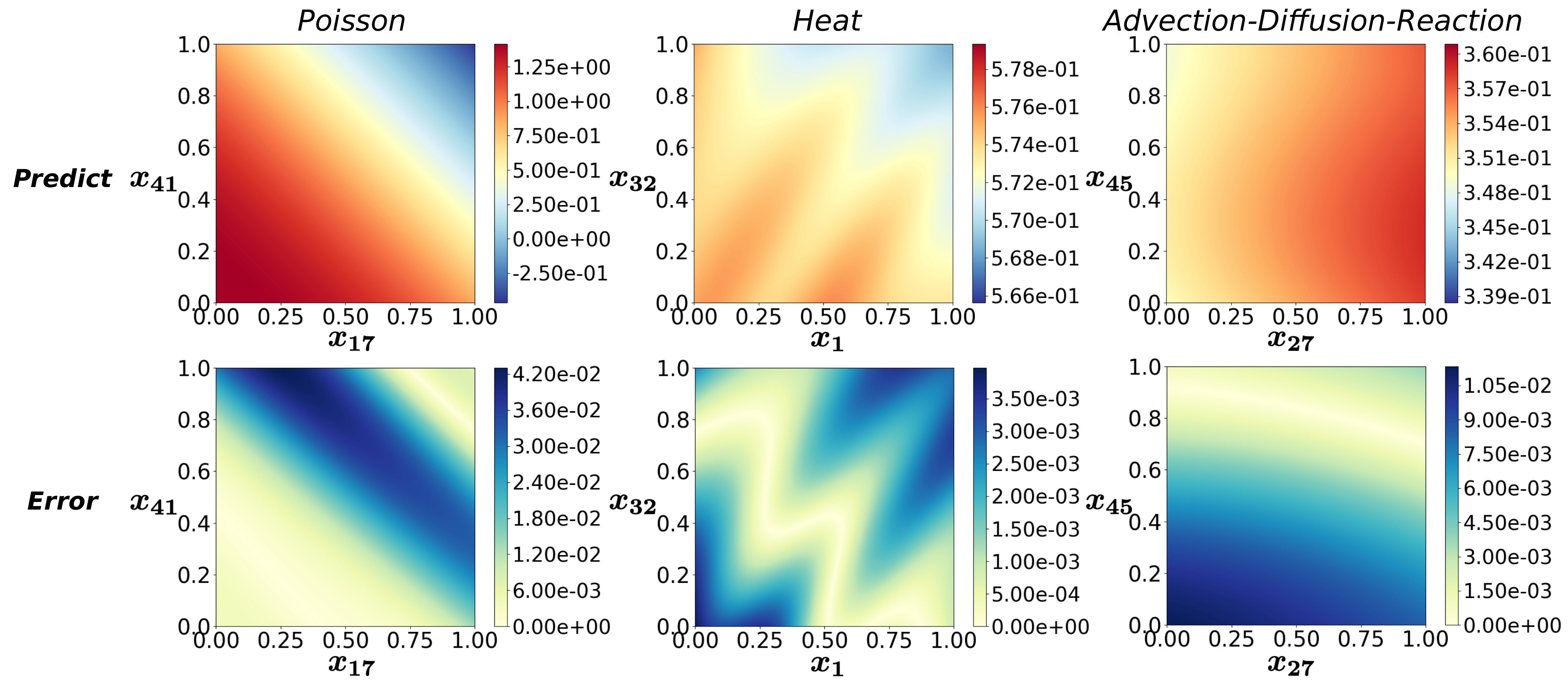}
    \caption{The figure presents contour maps of predicted values for high-dimensional, 50D advection-diffusion-reaction equation, 50D heat equation, and 50D Poisson equation, with two dimensions randomly selected for visualisation while fixing the remaining dimensions at random constant values within their domain. The map showcases the predicted values across the domain and the error compared to the true values for each equation, highlighting the model's ability to achieve accurate predictions in a high-dimensional context.}
    \label{fig:50_compare}
\end{figure}

\subsection{Discussions}
Through four carefully selected benchmark problems spanning different classes of PDEs,  PDE-DKL demonstrates promising success in mitigating the curse of dimensionality. Our experiments indicate that DNN-learned latent representations enable GP regression to maintain prediction accuracy (relative $L_2$ error < 5\%) even under limited training data. This synergy effectively addresses individual limitations, delivering accurate predictions and providing reliable uncertainty estimates. It highlights how DL’s computational power can be successfully combined with the theoretical rigour of GP approaches. The encouraging performance observed here suggests that future research could extend this framework to even more challenging PDEs, further leveraging advanced DL techniques for enhanced problem-solving in high-dimensional settings.

\section{Concluding remarks}
Machine learning models constrained by physics (including those based on NNs and GPs) often face significant challenges when tackling high-dimensional PDEs with sparse data. The curse of dimensionality typically demands a prohibitive amount of training data, which grows exponentially with the number of dimensions. In practice, such an extensive dataset is rarely feasible in real-world applications.
Our observations suggest that NNs and GPs should be regarded as complementary methods rather than alternatives in tackling these challenges. Conventional NNs excel at discovering low-dimensional manifolds in high-dimensional data but lack uncertainty quantification. By contrast, GPs can provide confidence estimates for each prediction, yet they struggle when scaling to extremely high dimensions.

To address these issues, we present a PDE-constrained statistical learning framework that synergistically combines GPs with DNNs for high-dimensional PDEs. The DNN component effectively identifies lower-dimensional manifolds, facilitating GP predictions in a reduced latent space and thus leveraging the complementary strengths of both. This synergy yields consistently accurate predictions while offering reliable uncertainty estimates, demonstrating robustness even under limited data scenarios. We have also enhanced the algorithm's efficiency and convergence through improved initialization and optimization strategies. Specifically, our method approximated three common 50-dimensional PDEs with high accuracy and rapid training.
Numerical experiments on several high-dimensional PDEs demonstrate that this unified framework not only improves predictive accuracy under limited training data but also broadens the scope of surrogate modelling in high-dimensional contexts. In particular, our PDE-DKL achieves higher accuracy and reliable uncertainty quantification than PDE-GP.

Some future research directions include refining the training dynamics of neural-network-based models for solving PDEs and exploring techniques such as random weight factorisation to accelerate convergence and enhance model performance. Additionally, the potential advantages of alternative activation functions, like sinusoidal functions (SIREN) and leaky ReLU, are under investigation. Incorporating curriculum training is also a part of our strategy to improve training efficiency. Further, conducting a detailed analysis of neural tangent kernel (NTK) matrices would be interesting in gaining a deeper understanding of convergence rates. Ultimately, our goal is to establish a transferable empirical recipe for model training applicable across various PDEs, thereby contributing to the robustness and efficiency of our approach.

\section*{Data accessibility}
The data and code supporting the findings of this paper are available on GitHub at the following repository: \href{https://github.com/VhaoYan/PDE-constrained-DKL}{https://github.com/VhaoYan/PDE-constrained-DKL}. 

\section*{Acknowledgements}
M.G. acknowledges the financial support from Sectorplan Bèta (the Netherlands) under the focus area \emph{Mathematics of Computational Science}. 
W.Y. acknowledges the China Scholarship Council for its support under No. 202107650017. 
M.G. and C.B. also acknowledge the support from the 4TU Applied Mathematics Institute for the Strategic Research Initiative \emph{Bridging Numerical Analysis and Machine Learning}.


\bibliographystyle{unsrtnat}
\bibliography{references}

\end{document}